\documentclass[11pt, oneside]{article}   	
\usepackage{geometry}                		
\usepackage{graphicx}				
\usepackage{amsmath,amssymb,epsfig,subfigure,mathtools,amsthm}

\theoremstyle{definition}

\theoremstyle{remark}

\usepackage{authblk}

\title{Enforcing constraints for interpolation and extrapolation in Generative Adversarial Networks}

\author[1]{Panos Stinis} 
\author[1]{Tobias Hagge}
\author[1]{Alexandre Tartakovsky} 
\author[2]{Enoch Yeung} 

\affil[1]{Advanced Computing, Mathematics and Data Division, Pacific Northwest National Laboratory, Richland WA 99354}
\affil[2]{Mechanical Engineering, University of California Santa Barbara, Santa Barbara CA 93106}
\date{}							

\begin{document}

\maketitle

\begin{abstract}
Generative Adversarial Networks (GANs) are becoming popular machine learning choices for training generators. At the same time there is a concerted effort in the machine learning community to expand the range of tasks in which learning can be applied as well as to utilize methods from other disciplines to accelerate learning. With this in mind, in the current work we suggest ways to enforce given constraints in the output of a GAN generator both for interpolation and extrapolation (prediction). For the case of dynamical systems, given a time series, we wish to train GAN generators that can be used to predict trajectories starting from a given initial condition. In this setting, the constraints can be in algebraic and/or differential form. Even though we are predominantly interested in the case of extrapolation, we will see that the tasks of interpolation and extrapolation are related. However, they need to be treated differently. 

For the case of interpolation, the incorporation of constraints is built into the training of the GAN. The incorporation of the constraints respects the primary game-theoretic setup of a GAN so it can be combined with existing algorithms. However, it can exacerbate the problem of instability during training that is well-known for GANs. We suggest adding small noise to the constraints as a simple remedy that has performed well in our numerical experiments. 

The case of extrapolation (prediction) is more involved. During training, the GAN generator learns to interpolate a noisy version of the data and we enforce the constraints. This approach has connections with model reduction that we can utilize to improve the efficiency and accuracy of the training. Depending on the form of the constraints, we may enforce them also during prediction through a projection step. We provide examples of linear and nonlinear systems of differential equations to illustrate the various constructions. 
\end{abstract}

\section{Introduction}\label{}
Advances in deep learning have transformed the modern outlook of artificial intelligence and machine learning.  Deep neural networks (DNNs) have demonstrated the ability to learn, generalize, and achieve state-of-the-art performance across a range of applications, including image and video classification, autonomous driving, robotic grasping, multi-player games, video captioning, natural language processing, translation and speech recognition (see e.g. the review \cite{lecunetal2015} and references therein).   

Deep learning so far has relied on its general advantageous properties which do not necessarily take into account specific laws obeyed by the data. However, there is a law in scientific computing (expressed in different ways e.g. in statistics it is called the Rao-Blackwell theorem \cite{liu2001}) which says the following: {\it any} property of the system under investigation (symmetry, conservation laws etc.) that is known should be built in the algorithm because it can result in significant gains in efficiency. In this spirit, instilling physical laws (constraints) to the deep learning algorithm or to the architecture can help. In the current work, we examine how training algorithms can incorporate constraints, thereby increasing the computational efficiency and fidelity of DNN models.

In particular, we assume that we are provided with time series data from a dynamical system whose equations we know. We want to use these time series data to train a DNN to represent the flow map of the dynamical system. This means that we want to train a DNN which can take as input the state of the system at a given time and output the state of the system after a certain time interval (timestep). We interpret the known equations of the dynamical system for which we have time series data as constraints that we want to enforce during training of the flow map DNN representation. Our aim is not to upend the  achievements of scientific computing but investigate the nexus of scientific computing and machine learning and how to take advantage of the two-way interaction between them (see also the recent DOE report on Scientific Machine Learning \cite{doe_sml_report}).    

To be more specific, in the current work we want to examine how enforcing directly a constraint in the context of a Generative Adversarial Network (GAN) can improve its efficiency and/or accuracy. We have chosen GANs as a framework to enforce constraints because of their promising performance in training generators and because they do not require supervision. We examine how to enforce constraints both for interpolation and extrapolation (prediction). The two cases need to be treated differently. 

For the case of interpolation,we have chosen to enforce the constraint by enhancing the input vector of the GAN discriminator to include the constraint residual i.e., how well a sample satisfies the constraint (see Section \ref{enforcing_constraints_interpolation}). The rationale behind this GAN variant is twofold: i) to introduce the constraint in a way that respects the game-theoretic setup of GANs and thus can be combined with existing algorithms; and ii) to introduce the functional relation of the constraint in the GAN value function so that it can be back-propagated to the generator. However, the incorporation of constraints can exacerbate the problem of instability during training that is well-known for GANs \cite{arjovskybottou2017}. We suggest adding small noise to the constraints as a simple remedy that has performed well in our numerical experiments (see Section \ref{add_noise}). Furthermore, we noticed that the GAN training algorithm can converge to its game-theoretic optimum before the actual interpolation error is brought to within an acceptable range. This signifies that there are narrow and deep crevices in the error landscape and we need an adaptive learning rate to reach them (see also the discussion in \cite{hodasstinis2018}). Based on this observation we have devised a learning rate scheme that can home in on these crevices (see Section \ref{learning_rate}).  

For the case of extrapolation, the enforcing of the constraints is more involved and an alternative approach is needed. To see why this is the case, let us suppose that we are given a trajectory (a time series) of a system and that the constraint we want to enforce are algebraic equations that the system variables satisfy at every point of the trajectory. For example, the system variables at each point of the trajectory may lie on a manifold. What we would like to learn is the map of the system i.e., how each point of the trajectory is mapped to the next one. Then, starting from an initial condition, we can apply this map iteratively to obtain a trajectory of the system. If we enforce the constraint during training, then the network learns how to interpolate a {\it single} trajectory. However, a single trajectory is extremely unlikely (has measure zero) in the phase space of the system. Thus, the trained network extrapolates accurately {\it as long as the extrapolated state remains on the training trajectory}. But when we extrapolate, every step involves an inevitable approximation error. If left unchecked, this approximation error causes the extrapolation to deviate into a region of phase space that the network has never trained on. Soon after, all the predictive ability of the network is lost. 

We suggest an alternative approach to training for extrapolation. In particular, we train the GAN generator to perform interpolation with {\it modified data}. We center a cloud of points at each data point that is provided on the trajectory. Then, during training, at each step we force the GAN generator to map a point in this cloud to the correct (noiseless) point on the trajectory at the next step. In other words, the GAN generator learns to interpolate {\it from this cloud of noisy data back to the given (noiseless) trajectory}. This construction is akin to introducing a restoring force in the dynamics of the system whose map the network has to learn. This restoring force is there to compensate for the inevitable approximation error committed at every step of the extrapolation process. In certain cases it can be interpreted as a memory term that appears in model reduction formalisms \cite{chorinstinis2007}. In addition to the introduction of a restoring force through the use of modified training data, one can choose to {\it also} enforce the given constraints during training. Depending on the form of the constraints and the form of the generator output, the restoring force can appear explicitly in the constraints (the various possibilities are clarified in the numerical examples in Section \ref{instructive_examples}). 

After training, we use the GAN generator to perform extrapolation (prediction). Depending on the form of the constraints e.g. manifold constraints, we may enforce them {\it also} during prediction through a projection step. The projection step acts as a correction step which forces the GAN output to conform to the constraints. We emphasize that this projection step is not possible for every system but may also not be necessary. We provide numerical examples to illustrate the various possible constructions.

In recent years, there has been considerable interest in the development of methods that utilize data and physical constraints in order to train predictors for dynamical systems and differential equations e.g. see \cite{PhysRevE.91.032915, raissi2018, chen2018, Han8505, SIRIGNANO20181339, felsberger2018, wan2018, MaE9994} and references therein. Our approach is different, it introduces the novel concept of training on purpose with modified (noisy) data in order to incorporate a restoring force in the dynamics learned by the generator. We have also provided the connection between the incorporation of such restoring forces and the concept of memory in model reduction. 

The paper is organized as follows. Section \ref{constraints_gans} contains the basic GAN framework (Section \ref{gan_framework}) as well as the proposed way to enforce constraints both in the case of interpolation (Section \ref{enforcing_constraints_interpolation}) and extrapolation (Section \ref{enforcing_constraints_extrapolation}). Section \ref{add_noise} describes how we decide on the magnitude of the noise to be added to the constraints for the true data in order to tame the well-known GAN instability. Section \ref{learning_rate} describes a learning rate schedule that we have devised to aid the convergence of a GAN. Section \ref{instructive_examples} contains numerical results. Section \ref{discussion} contains a brief discussion of the constructions and the results as well as several suggestions for future work.


\section{Constraints and GANs}\label{constraints_gans}

Suppose that we are given a dynamical system described by an M-dimensional set of differential equations 

\begin{equation}\label{odes}
\frac{dx}{dt}=f(x),
\end{equation} 
where $x \in \mathbb{R}^M.$ The system \eqref{odes} needs to be supplemented with an initial condition $x(0)=x_0.$ Furthermore, suppose that we are provided with time series data from the system \eqref{odes}. This means a sequence of points from a trajectory of the system $\{x_i^{data}\}_{i=1}^N$ recorded at time intervals of length $\Delta t.$ We would like to use this time series data to train a neural network to represent the flow map of the system i.e. a map $H^{\Delta t}$ with the property $H^{\Delta t}x(t)=x(t+\Delta t)$ for all $t$ and $x(t).$ 

We want to find ways to enforce {\it during training} the constraints implied by the system \eqref{odes}. In addition to \eqref{odes}, one could have extra constraints. For example, if the system \eqref{odes} is Hamiltonian, we have extra algebraic constraints since the system must evolve on an energy surface determined by its initial condition. The framework we present below allows the enforcing of both differential and algebraic constraints.

While it is desirable to enforce known constraints about the data for different DNN architectures, here we will focus on enforcing constraints for GANs, which constitute a recently introduced and popular method for unsupervised learning \cite{goodfellowetal2014}. The main reason for the popularity of GANs is their game-theoretic reformulation of the problem of training a generator to produce data from a given distribution (called the true data distribution). The adversaries of this game are the generator and an associated discriminator. The objective of the generator is to trick the discriminator into deciding that the generator-created samples actually come from the true data distribution. In order to achieve this, the discriminator needs to be trained with two types of samples: i) samples from the true distribution and ii) samples created by the generator. The generator and the discriminator are trained in tandem. The best possible performance of the generator is to convince the discriminator {\it half of the time} that the samples it is generating come from the true distribution. In such a case, the discriminator's ability to discriminate between true and generator-created samples is equivalent to deciding based on the toss of a fair coin.  

In Section \ref{gan_framework} we present the basic GAN framework. In Section \ref{enforcing_constraints_interpolation} we discuss how we can enforce constraints in GAN for {\it interpolation}. This involves augmenting the input of the discriminator for each sample by the constraint residual i.e. how well the sample satisfies the constraint. This applies to both the GAN generated samples as well as those coming from the true data distribution. In Section \ref{enforcing_constraints_extrapolation} we discuss how we can enforce constraints in GAN for {\it extrapolation}. Section \ref{add_noise} discusses our proposal to add noise to the constraint residual for the true data distribution samples in order to promote stability. Finally, Section \ref{learning_rate} discusses a custom-made learning rate schedule for training GANs. 

\subsection{The basic GAN framework}\label{gan_framework}
Generative Adversarial Networks comprise of two networks, a generator and a discriminator. The target is to train the generator's output distribution $p_g(x)$ to be close to that of the true data $p_{data}.$ We define a prior input $p_z(z)$ on the generator input variables $z$ and a mapping $G(z;\theta_g)$ to the data space where $G$ is a differentiable function represented by a multilayer perceptron with parameters $\theta_g.$ We also define, a second multilayer perceptron (the discriminator) $D(x;\theta_d)$ which outputs the probability that $x$ came from the true data distribution $p_{data}$ rather than $p_g.$ We train $D$ to {\it maximize} the probability of assigning the correct label to both training examples and samples from the generator $G.$ Simultaneously, we train the $G$ to minimize $\log (1-D(G(z))).$ We can express the adversarial nature of the relation between $D$ and $G$ as the two-player minimax game with value function $V(D,G)$:
\begin{equation}\label{game_gan}
\min_G \max_D V(D,G) = E_{x \sim p_{data}(x)}[\log D(x)] +E_{z \sim p_z(z)}[\log (1- D(G(z)))].
\end{equation}
There are certain properties satisfied by the game-theoretic setup of training in GANs \cite{goodfellowetal2014}. For a given generator $G,$ the optimal discriminator $D_G^*(x)$ is given by 
$$D_G^*(x) = \frac{p_{data}(x)}{p_{data}(x)+p_g(x)}$$
We can use $D_G^*(x)$ to define an objective function for the training of the generator $G.$ In particular, we can define $C(G)=\max_D V(D,G)=V(D_G^*,G).$ Then, one can show that the global minimum of $C(G)$ is obtained if and only if $p_g=p_{data}.$ Also, since for this minimum we have $D_G^*(x) =\frac{1}{2},$ the value of the minimum is $- \log 4.$ Correspondingly, for the generator, the minimum of $\log (1-D(G(z)))$ is $- \log 2.$ 

These theoretical properties allow us to monitor the progress of the training in its game-theoretic sense. However, there are two important issues that make the behavior of GANs in practice delicate. First, the generator is usually weaker than the discriminator, so that the convergence to the game-theoretical optimum is not guaranteed \cite{arjovskybottou2017}. Second, even if there is convergence to the game-theoretical optimum there is no guarantee that this optimum corresponds to a well-trained generator \cite{arjovskybottou2017}. These issues have resulted in an increasing literature of variants of the original GAN formulation as well as suggestions and guidelines about the training procedure. In the current work, we have implemented the simple GAN framework as presented in \cite{goodfellowetal2014}, with the modification suggested there to replace the minimization of $\log (1-D(G(z)))$ by the minimization of $-\log (D(G(z))).$  

Another issue that one has to address is the choice of the learning rate during training e.g. when Stochastic Gradient Descent is used. A naive choice of the learning rate can lead to instability of the GAN training. Looking at the definition of the objective functions to be optimized, instability manifests as unbounded growth of  $-\log (D(G(z)))$ and convergence of $\log (1-D(G(z)))$ to 0. While our main focus in the current work is the enforcement of constraints we have also devised a scheme to choose the learning rate in order to avoid such instabilities (see Section \ref{learning_rate}). The scheme we have devised is based on general observations about the learning pattern of DNNs and has performed satisfactorily in the numerical experiments. Still, the decisive factor in the acceleration of training of a GAN to enforce a constraint is {\it whether} this constraint is incorporated in the GAN setup as we will see in the next section.

\subsection{Enforcing constraints for {\it interpolation} with GANs}\label{enforcing_constraints_interpolation}
We are interested in training the generator $G$ to represent the flow map of the dynamical system. That means that if $z$ is the state of the system at a time instant $t,$ we would like to train the generator $G$ to produce as output $G(z),$ an accurate estimate of the state of the system at time $t+ \Delta t.$ In addition, we want to enforce constraints in the output of the generator $G$ in a way that respects the game-theoretic setup of GANs. We can do so by {\it augmenting} the input of the discriminator by the constraint residuals i.e. how well does a sample satisfy the constraints. Of course, such an augmentation of the discriminator input should be applied {\it both} to the generator-created samples as well as the samples from the true distribution. This means that we consider a two-player min-max game with the modified value function $V^{constraints}(D,G):$
\begin{gather}
\min_G \max_D V^{constraints}(D,G) \notag \\
= E_{x \sim p_{data}(x)}[\log D(x,\epsilon_D(x))] +E_{z \sim p_z(z)}[\log (1- D(G(z),\epsilon_G(z)))],\label{game_gan_constraints}
\end{gather}     
where $\epsilon_D(x)$ is the constraint residual for the true sample and $\epsilon_G(z)$ is the constraint residual for the generator-created sample. Note that in our setup, the generator input distribution $p_z(z)$ will be from the noise cloud around the training trajectory. On the other hand, the true data distribution $p_{data}$ is the distribution of values of the (noiseless) training trajectory.

The constraint residuals $\epsilon_D(x)$ and $\epsilon_G(z)$ for the true and generator-created samples, respectively, measure how well the samples, true and GAN generated, satisfy the constraints. The constraints can be at the different levels of description of the dynamical system e.g. \eqref{odes} or at an algebraic level e.g. after numerical discretization (please see \eqref{constraint_intra_residual_1} in Section \ref{example_interpolation_explicit}, \eqref{constraint_intra_residual_2} in Section \ref{example_interpolation_implicit}, \eqref{constraint_extra_1_residual}-\eqref{constraint_extra_2_residual} in Section \ref{example_extrapolation_linear} and \eqref{lorenz_modified1_residual}-\eqref{lorenz_modified3_residual} in Section \ref{example_extrapolation_nonlinear} for concrete examples).

The samples from the true distribution satisfy these constraints identically and thus $\epsilon_D(x)$ can be set to zero. However, as we have already mentioned in the introduction and also discuss in detail in Section \ref{add_noise} below, this can hinder the training process by giving the, already favored by the GAN setup, discriminator an even larger advantage over the generator. The situation can be remedied by turning $\epsilon_D(x)$ into a random variable with zero mean and small variance i.e. by adding noise to the constraint residuals for the true samples. In order not to digress from our discussion about enforcing constraints we have postponed the discussion about the choice of the variance of the noise until Section \ref{add_noise}.  

On the other hand, the GAN generator needs to train in order to produce samples that satisfy the constraints. Thus, $\epsilon_G(z)$ is an expression which measures how well a GAN generated sample satisfies the constraint. The advantage of our formulation is that $\epsilon_G(z)$ contains the functional form of the constraint as a function of the GAN generator parameters i.e. the weights and biases of the neural network that represents the generator. In this way, when using back-propagation, in the GAN generator update step, we differentiate through the functional form of the constraint. In this way, the actual functional form of the constraint becomes part of the GAN generator training. We should note that the implementation of our formulation is straightforward with contemporary machine learning tools like TensorFlow \cite{abadietal2016}.

\subsection{Enforcing constraints for {\it extrapolation} with GANs}\label{enforcing_constraints_extrapolation}

The training of a GAN to perform extrapolation is different from that for interpolation. Extrapolation requires the iterative application of the GAN generator (in a feedback loop) starting from an initial condition so as to produce a trajectory of a system. To train the GAN generator we assume that we have data from the system in the form of a time series. These data are used by the GAN generator to learn the map of the system i.e., how to produce the next point on a trajectory of the system given one point.

There is a fundamental difference between interpolation and extrapolation. In interpolation, we have a collection of values of the unknown function at different points and we wish to produce values of the function for points in between the ones given. Even though there is inevitable error for the interpolation between two points, this error concerns only the interval between these two points. On the other hand, in the case of extrapolation, the error committed in one step propagates to the next step where a new error is committed and so on. This compounding of errors can quickly lead the extrapolation astray.

When we enforce the constraints during the training of the GAN generator to be used for extrapolation we are {\it restricting} the area of phase space that is available to the generator to train on. When we are provided with a trajectory of the system and we enforce during training the fact that the data came from this specific trajectory, we are restricting the generator to train {\it only} on this trajectory. But a trajectory has measure zero in the space of all possible trajectories. As a result, during extrapolation, the inevitable errors committed during each step will send the generator output to parts of the phase space that the generator has never been exposed to before. This situation suggests that a different approach is required for the training of a generator in order to become an accurate extrapolator.

\subsubsection{Training with noisy data to promote the extrapolation accuracy}
In this section we propose the idea of adding noise to the training data in order to produce a GAN generator with improved extrapolation accuracy. The use of noisy data can be combined with the enforcing of constraints. Moreover, the enforcing of constraints can be done i) only during training or ii) both during training and prediction. This is a problem-dependent decision and in Section \ref{instructive_examples} we provide numerical examples for both choices. 

\subsubsection*{Enforcing constraints {\it only} during training}

We have said that the extrapolation task depends on the accuracy with which the GAN generator learns the map of the system. But learning the map of the system is learning an interpolation task which has some built-in error. This means that the trained GAN generator at each extrapolation step will commit an error. Thus, if we hope to train an accurate extrapolator we should find a way to compensate for this error. We would like to train the GAN generator to ``correct" its trajectory after every step so that it counters the effect of the error. 

Here we propose to train the generator on a ``noisy" version of the given trajectory. We center a cloud of points at each data point that is provided on the trajectory. Then, during training, at each step we force the generator to map a point in this cloud to the correct (noiseless) point on the trajectory at the next step. The purpose of this is for the generator to learn how to restore the trajectory of the system after it is perturbed. The proposed approach introduces two new parameters that need to be determined in the numerical experiments: i) the range in which the cloud of points takes values and ii) how many points will the cloud consist of. In the current paper we provide only empirical criteria to tune these parameters. Since we have limited computational capability we wish to achieve a balance between the range of the noise cloud and the number of points that we will sample from the cloud. If we use a small range, then a few points will suffice. However, in this case we don't explore enough the nearby points of the given trajectory. On the other hand, if we use a rather large range we may not have enough points to sample it adequately and this can degrade the accuracy of the trained network. Similar trade-offs can be found in recent work on bounds on the generalization capability of neural networks (see e.g. \cite{xu2017, bu2019, kunze2019}).

The system dynamics learned by the GAN generator, in lapidary form ``GAN generator dynamics = approximate original dynamics + restoring force", hint at connections between our approach and model reduction \cite{chorinstinis2007}. In particular, the restoring force is analogous to the concept of memory in the context of model reduction. Whether we need to decompose {\it explicitly} the dynamics learned by the GAN generator into a sum of approximate original dynamics and a restoring force or the restoring force can be incorporated {\it implicitly} in the GAN generator dynamics depends on the type of constraint that we want to enforce and the level at which we examine the dynamical system e.g. state space vs phase space (the various cases are clarified through the numerical examples in Section \ref{instructive_examples}). 

For the example of the two coupled linear oscillators (see Section \ref{example_extrapolation_linear}), we examine the system at the level of the phase space. We enforce the restoring force in the GAN generator dynamics implicitly. For the Lorenz system (see Section \ref{example_extrapolation_nonlinear}), we examine the system at the level of the state space. We have explicitly enforced during training the constraint that the pair of input and output of the GAN generator satisfies an Euler scheme modified by the addition of a simple linear memory term. We have found that even this simple linear memory can improve significantly the extrapolation accuracy of the GAN generator. More elaborate memory terms and a more thorough investigation of the this connection will appear in a future publication.

\subsubsection*{Enforcing constraints during training {\it and} prediction}

We can enforce constraints both during training and prediction (see Section \ref{example_extrapolation_linear}). Of course, whether this is feasible and advisable is case dependent. We note that there is a significant difference in enforcing a constraint during training and enforcing a constraint through a projection step during extrapolation. To enforce a constraint during training, we need to augment the input vector of the GAN discriminator with the constraint residual as explained in Section \ref{enforcing_constraints_interpolation}. To enforce a constraint through a projection step during extrapolation one has to {\it find a solution that satisfies the constraint}. For nonlinear systems this task can be turned into an optimization problem which, in general, can be costly. However, some of the optimization cost can be avoided if we initialize the optimizer with the output of the GAN generator (before the projection).

\subsection{Adding noise to the constraint residual to promote stability}\label{add_noise}   
We have seen that whether we train a GAN for interpolation or extrapolation purposes, a constraint can be incorporated directly. However, there is a delicate matter that needs to be addressed in order to harvest the advantages of introducing directly the constraint in the GAN framework. As we have mentioned before, the practice of training GANs has shown that the discriminator can train faster and more effectively than the generator. This is to be expected especially if one takes into account that the job of the discriminator amounts to coarse-graining while that of the generator to refinement \cite{mehtaschwab2014, linetal2017}. When we augment the training of the discriminator with the introduction of the constraint residual $\epsilon$ we can accelerate the discriminator's training to a point where it becomes too strict. What this means is that although the generator's training is {\it also} accelerated, it cannot catch up with the discriminator. As a result, even though the generator can produce fine samples that satisfy the constraint to a high accuracy, they are almost all rejected by the discriminator. This causes the GAN training to become unstable. Such a situation of the generator producing good samples yet the GAN training suffering from instability has been documented for GANs in a different context \cite{arjovskybottou2017}.

To avoid the instability we can regulate the discriminator's ability to distinguish between generator-created and true data by adding noise to the constraint residual {\it for the true data only}. The idea behind adding noise is that the discriminator will become more tolerant by allowing generator-created samples for which the constraint residual is not exactly 0 but essentially within a narrow interval around it. 

The magnitude of the noise can be based on different considerations depending on the type of constraints that we want to enforce. For the numerical example in Section \ref{example_extrapolation_linear}, the magnitude of the noise was chosen based on general Monte-Carlo considerations \cite{liu2001} and prior experience with stochastic optimization algorithms \cite{stinis2012}. In particular, if we are using $M$ samples in our training set, then we can add to the constraint residual for each sample a random push $\sim {\cal{N}}(0,c^2/M)$ where $c$ is a user-chosen parameter i.e. a sample from a Gaussian with mean 0 and standard deviation  $c/\sqrt{M}.$ For the numerical example in Section \ref{example_extrapolation_nonlinear}, the magnitude of the noise was based on the numerical accuracy of the scheme (the Euler scheme) that was used to produce the ground truth.

\subsection{Selection scheme for the learning rate}\label{learning_rate}
The practice of training GANs with Stochastic Gradient Descent (SGD) has shown that one has to choose carefully the learning rate to avoid slow convergence and possible instabilities (we have explained before the meaning of instability in the GAN context). As will be also apparent from the numerical examples, in order to gauge the progress of training we should use quantities that are different from the convergence to the game-theoretic optimum. Based on general observations about the learning pattern of deep neural networks, we have devised a scheme for selecting the learning rate during training. 

In the numerical experiments we monitor the progress of the GAN's ability to learn a constraint. In particular, we monitor the evolution (as the SGD iterations proceed) of the relative error over a mini-batch of size $m$   
$$ RE_m= \frac{1}{m}\sum_{j=1}^m \frac{|G(z_j)-f(z_j)|}{|f(z_j)|},$$
where $G(z_j)$ is the output of the GAN generator for the input $z_j$ and $f(z_j)$ the true value for the same input. We should note here that, even though we monitor the relative error to decide the learning rate, our algorithm remains {\it unsupervised} since this information is {\it not} back-propagated through the network in order to update the weights.

Our main observation about the behavior of the training process is that the neural network training proceeds in a succession of sharp drops in the value of $RE_m$ followed by ``quiet" periods where the optimizer explores the weight space without losing accuracy i.e., exhibiting small oscillations in the value of $RE_m.$  Based on that we have decided to check for the need to adjust the learning rate every $N_{check}$ iterations. In other words, we let the training proceed by dividing it in parcels, each containing $N_{check}$ iterations. In the numerical experiments we chose $N_{check}=2000.$ We collect the values of $RE_m$ for the $N_{check}$ iterations and then we decide whether it is time to adjust the learning rate. In order to decide we have encoded our observations about the general pattern of training into 3 checks. If any of the 3 checks is true, then we reduce the learning rate by a factor $\alpha.$ The 3 checks are:
\begin{itemize}
\item Compute the {\it minimum} of the relative error over the $N_{check}$ iterations and check whether it lies in the $(r_{lower},r_{upper})$ range of iterations,
\item Check whether the value of the relative error at the $N_{check}$th iteration is {\it larger} than the value at the first iteration,
\item Check whether the reduction of value of the relative error between the first and $N_{check}$th iterations is {\it smaller} than $r_{drop}$ of the value of the relative error of the first iteration.
\end{itemize} 
The parameters appearing in the 3 checks were set to the values $r_{lower}=20\%,$ $r_{upper}=80\%$ and $r_{drop}=25\%$ for the numerical experiments. Also, the learning rate reduction factor was taken to be $\alpha=2.$ We note here that these values are not optimized. They are based on general observations about how the neural network training proceeds.

Finally, we need to provide some criterion to decide when to terminate the training of the GAN. We implemented two termination criteria, one to address convergence and the other the lack of convergence. Since the algorithm is stochastic we postulated that the algorithm has converged if $RE_m \leq TOL$ for at least {\it half} of the $N_{check}$ iterations within an iteration parcel. To address the lack of convergence we set a {\it minimum} allowed value of the learning rate. For the numerical experiments this minimum was taken to be $10^{-12}.$


\section{Numerical results}\label{instructive_examples}
We present numerical results for different cases in order to exhibit the significant increase in efficiency and/or accuracy that can be achieved in training a GAN when we enforce directly a constraint. We have to clarify that for the experiments where we do {\it not} enforce the constraints, we still augment the discriminator input vector by the generator input in addition to the generator's output. On the other hand, when we {\it do} enforce the constraints, we further augment the discrimination input vector by the constraint residuals in addition to the pair of generator input and output vectors. Thus, what we refer to as the case of {\it not enforcing constraints} is different from the vanilla use of GANs where the only information used from the generator as input for the discriminator is the generator's output. Of course, similar adjustments are afforded to the samples from the true data (recall the discussion in Section \ref{enforcing_constraints_interpolation}).  

We want to emphasize here that all the results presented correspond to the typical behavior observed over a large number of experiments. Obviously, since the training algorithm is random there is bound to be variance in the performance of each approach. Respecting this inevitability we did {\it not} handpick the results of a certain approach to make a point but presented results that exhibit the overall trends found in the experiments.  

\subsection{Enforcing constraints for interpolation}\label{example_interpolation}
For the case of interpolation we present two examples: i) a one-dimensional example with an explicit constraint (Section \ref{example_interpolation_explicit}) and ii) a two-dimensional example with an implicit constraint (Section \ref{example_interpolation_implicit}).

\subsubsection{One-dimensional example with explicit constraint}\label{example_interpolation_explicit}
Suppose that we are given $z \sim U[0,1]$ and we want to train the generator of a GAN to produce samples that satisfy $x=1-(2z-1)^2.$ 

Both the generator and discriminator are modeled as convolutional neural networks with an input, an output and 5 hidden layers each of width 10. All the hidden layers have as activation functions Exponential Linear Units (ELUs) \cite{clevertetal2015}. The only layer that is different is the last layer of the discriminator whose output is a sigmoid function because $D(x)$ and $D(G(z))$ are probabilities. For the numerical experiments we have used $M=10^4$ samples (M/3 used for training, M/3 for validation and M/3 for testing) and mini-batch SGD was used with a mini-batch size of $m=10^3.$ The relative error criterion for the learning rate choosing scheme is set to $TOL = 3/\sqrt{M/3}=3 /\sqrt{10^4/3} \approx 0.052.$ The reason we chose the tolerance to be equal to $3 \times \frac{1}{\sqrt{M/3}}$ is because we assume that the error is Gaussianly distributed with zero mean and standard deviation $\frac{1}{\sqrt{M/3}}.$ So, we allow the threshold to be equal to 3 standard deviations which is the range within which $99.8\%$ of a Gaussian random variable's mass lies. 

The constraint residual for the GAN generator output is given by 
\begin{equation} \label{constraint_intra_residual_1}
\epsilon_G(z)= G(z) - (1-(2z-1)^2). 
\end{equation}
The constraint residual for the true samples was taken to be $\epsilon_D(x) \sim {\cal{N}} (0,0.01^2/M)$ which corresponds to a Gaussian random variable with mean 0 and standard deviation $0.01/\sqrt{M}=0.01/100.$ As we have already mentioned in Section \ref{add_noise} this choice was guided from previous experience with stochastic optimization algorithms and general MC considerations. In general, we do not want to add a noise that is too large because that would defeat the purpose of enforcing the constraint.

\begin{figure}[ht]
   \centering
   \subfigure[]{%
   \includegraphics[width = 7cm]{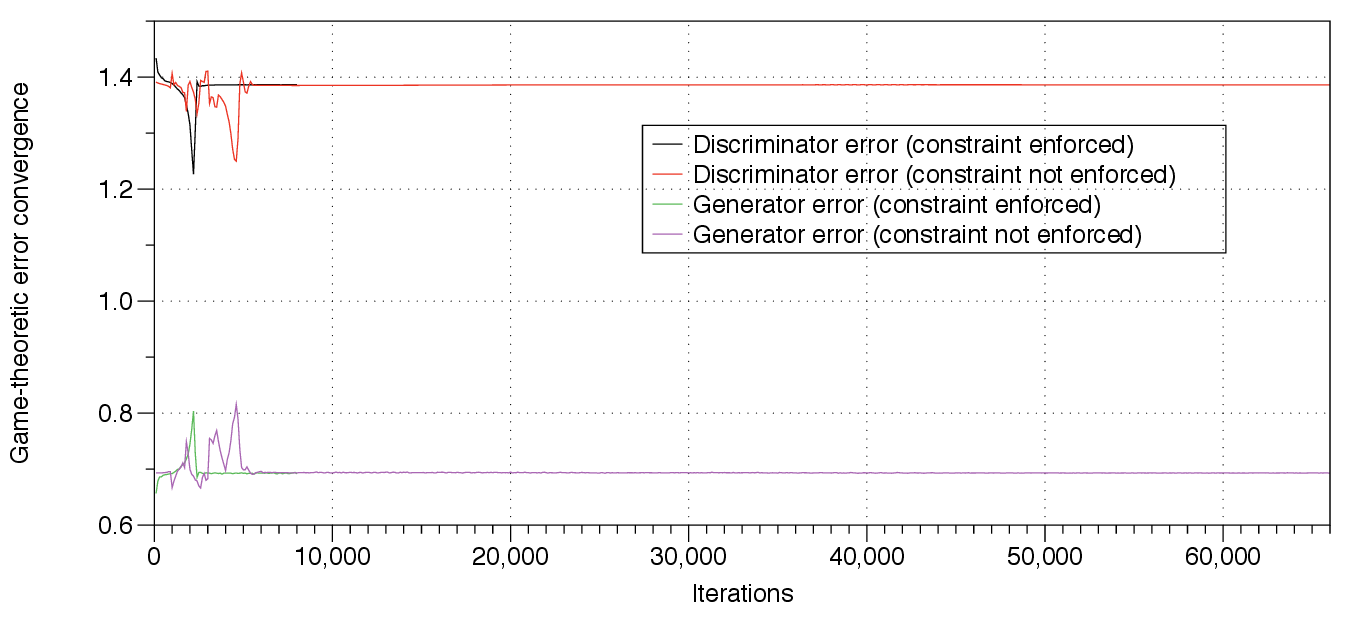}
   \label{plot_symmetric_1a}}
   \quad
   \subfigure[]{%
   \includegraphics[width = 7cm]{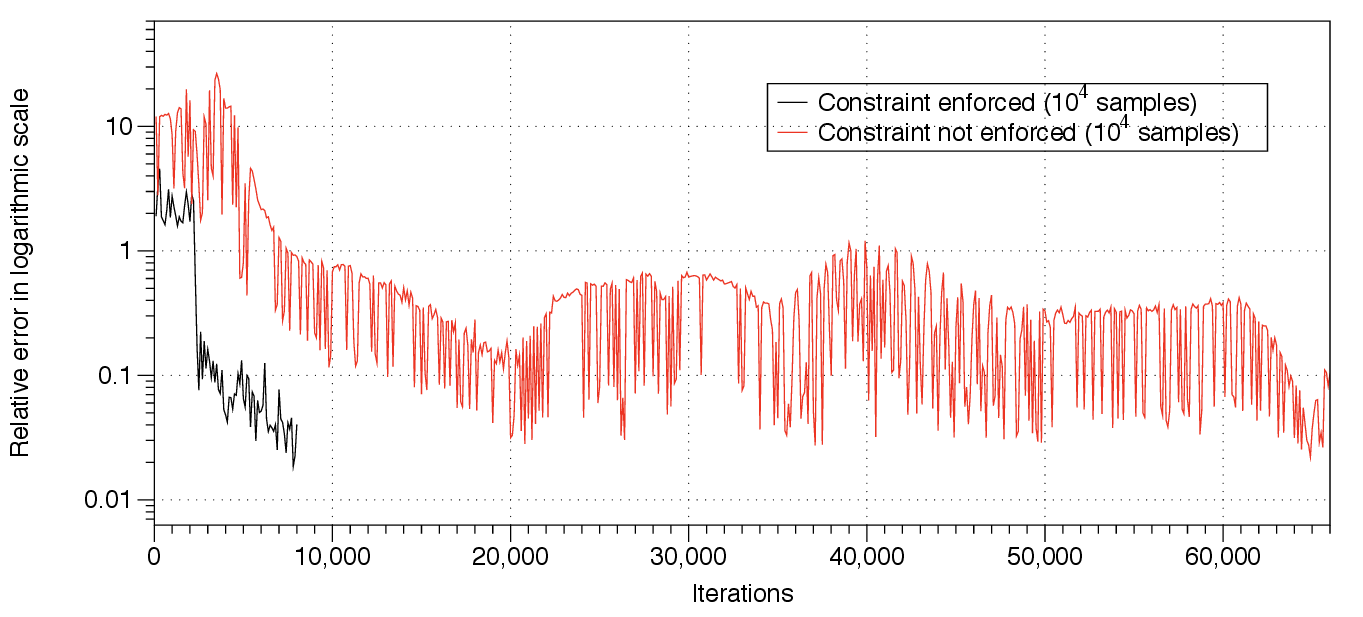}
   \label{plot_symmetric_1b}}
\caption{(a) Comparison of the evolution of the absolute value of the generator and discriminator game-theoretic error with and without enforcing a constraint (linear-linear plot) ; (b) Comparison of the evolution of the relative error $RE_m $ of the function learned with and without enforcing a constraint (linear-log plot). }
\label{plot_symmetric_1}
\end{figure}

Following the discussion in Section \ref{gan_framework} we include results from different diagnostics to ensure that the generator does indeed produce samples $x$ that satisfy the prescribed constraint. Fig. \ref{plot_symmetric_1a} shows results for the evolution of the {\it absolute value} of the game-theoretic error towards the optimum value. In particular, the {\it absolute value} of the discriminator error should converge to $\ln 4 \approx 1.3863$ while the {\it absolute value} of the generator error should converge to $\ln 2 \approx 0.6931.$ Whether or not we enforce the constraint the absolute values of the errors for the discriminator and the generator converge to about 1.3861 and 0.6934 respectively. We can translate these values to the relative errors of the learned probabilities $D(x)$ and $D(G(z))$ from their theoretical values. We find that the relative error for the discriminator error is about $0.0001\%$ while for the generator is $0.01\%$. As expected from the construction and theoretical properties, the discriminator trains more efficiently than the generator. Both come very close to the game-theoretic optimum. From Fig. \ref{plot_symmetric_1a} we see that this convergence is achieved early on. However, this is not enough to guarantee that the target function $x=1-(2z-1)^2$ is actually learned. 

\begin{figure}[ht]
   \centering
   \subfigure[]{%
   \includegraphics[width = 7cm]{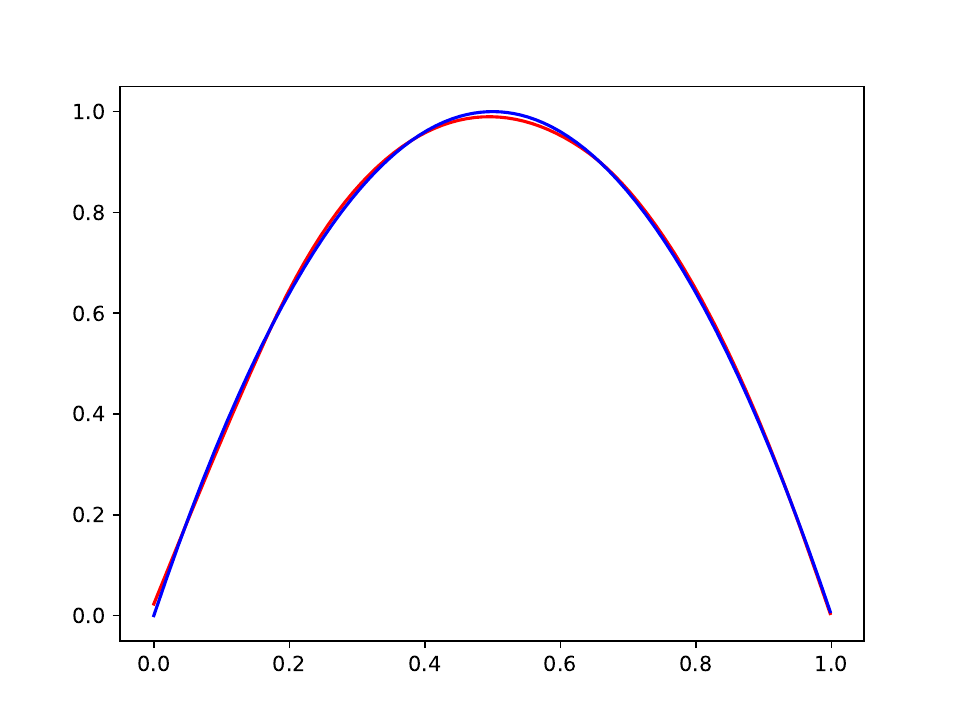}
   \label{plot_symmetric_2a}}
      \quad
   \subfigure[]{%
   \includegraphics[width = 7cm]{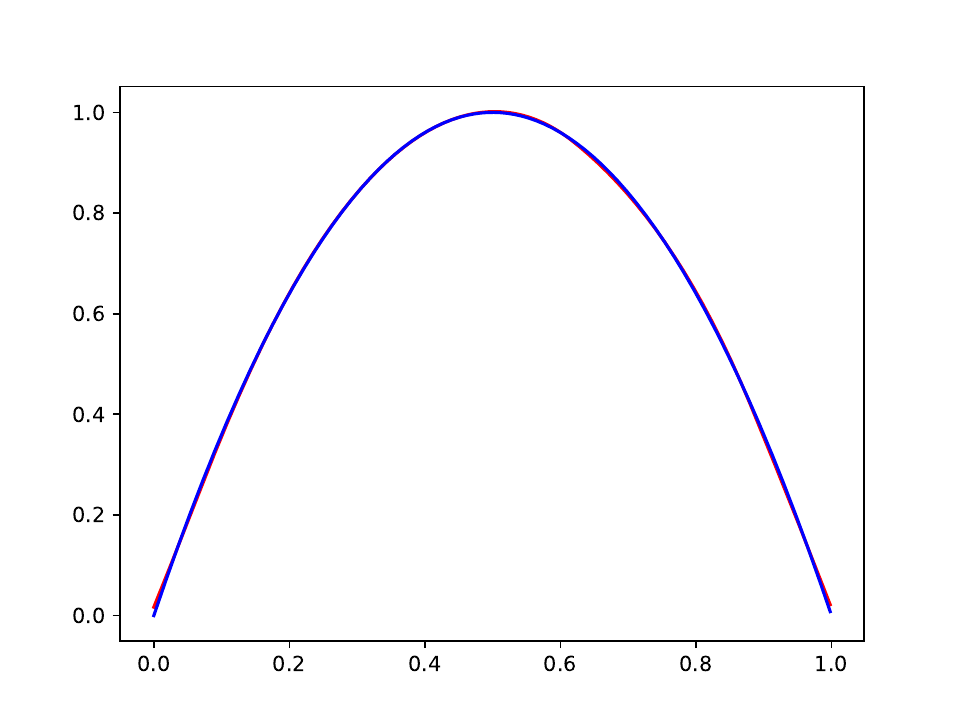}
   \label{plot_symmetric_2b}}
\caption{Comparison of the target function $x=1-(2z-1)^2$ (blue line) and actual function learned (red line). (a) $10^4$ samples without enforcing the constraint; (b) $10^4$ samples with the constraint enforced.  }
\label{plot_symmetric_2}
\end{figure}

To measure how well the target function is learned we monitor in Fig. \ref{plot_symmetric_1b} the relative error of the actual function that is learned by the generator. This relative error over the mini-batch of size $m$ is computed as  
$$ RE_m=\frac{1}{m}\sum_{j=1}^m \frac{|G(z_j)-(1-(2z_j-1)^2)|}{|1-(2z_j-1)^2|} ,$$
where $G(z_j)$ is the output of the GAN generator for the input $z_j.$ From Fig. \ref{plot_symmetric_1b}, we see that if we enforce the constraint in the GAN setup we achieve convergence to within the relative error criterion $TOL$ after 8000 iterations. On the other hand, if we do {\it not} enforce the constraint we need about 60000 iterations to achieve convergence. From Fig. \ref{plot_symmetric_2} we see that the actual function learned is very close to the target function whether we enforce the constraint or not. However, enforcing the constraint leads to a significant increase in the efficiency of the training.

\subsubsection{Two-dimensional example with implicit constraint}\label{example_interpolation_implicit}
For this example we consider the prescribed evolution of a system with two components which satisfy an algebraic constraint at every instant. This example is a caricature of evolution of a system on a surface e.g. a Hamiltonian system which evolves on the energy surface defined by the constant value of the Hamiltonian. This is a more involved example than the previous for two reasons: i) there are now two functions to be learned instead of one and ii) the constraint does not dictate directly the form of the two functions but only implicitly. 

\begin{figure}[ht]
   \centering
   \subfigure[]{%
   \includegraphics[width = 7cm]{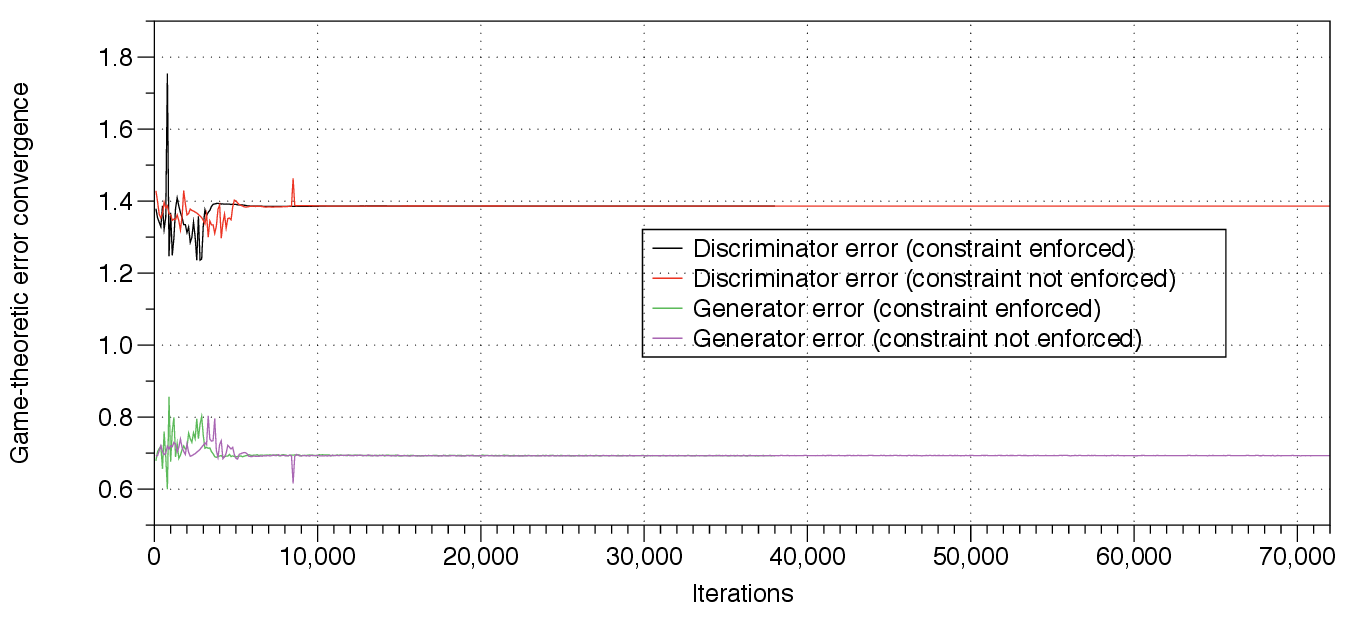}
   \label{plot_harmonic_1a}}
   \quad
   \subfigure[]{%
   \includegraphics[width = 7cm]{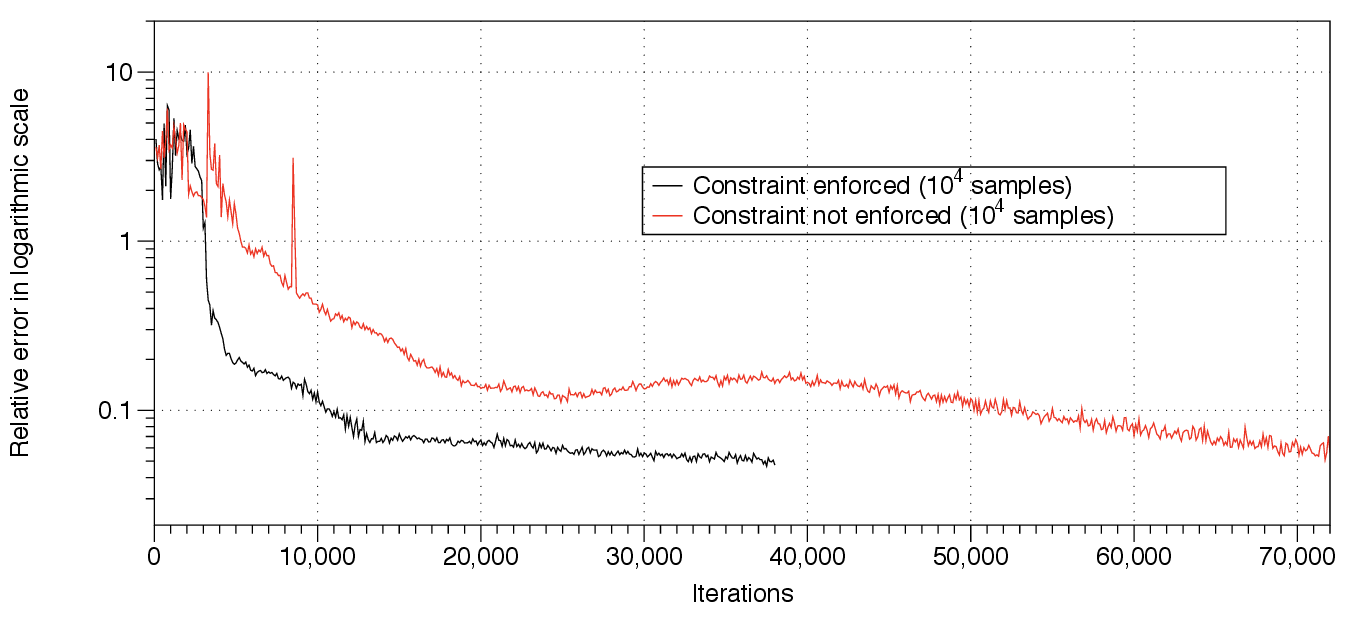}
   \label{plot_harmonic_1b}}
\caption{Enforcing a constraint in higher dimensions. (a) Comparison of the evolution of the absolute value of the generator and discriminator game-theoretic error with and without enforcing a constraint (linear-linear plot) ; (b) Comparison of the evolution of the relative error $RE_m $ of the function learned with and without enforcing a constraint (linear-log plot).  }
\label{plot_harmonic_1}
\end{figure}

Suppose that we are given the functions 
\begin{gather}
x_1(z)=R-\sin(z) \label{evolution_1}\\
x_2(z)=R-\cos(z) \label{evolution_2}
\end{gather}   
where $R$ is a prescribed constant and $z$ takes values in a prescribed interval. One can consider the evolution in \eqref{evolution_1}-\eqref{evolution_2} as the solution of the simple system of ordinary differential equations
\begin{gather}
\frac{d}{dz}(x_1-R)=x_2-R \label{ode_1}\\
\frac{d}{dz}(x_2-R)= -(x_1-R).  \label{ode_2}
\end{gather}   
It is easy to see that 
\begin{equation}
(x_1(z)-R)^2+(x_2(z)-R)^2=1 \;\; \text{for all} \; \; z, \label{ode_constraint}
\end{equation}
which means that the point $(x_1(z),x_2(z))$ stays on the unit circle centered at $(R,R).$ We want to train a GAN to produce samples $G(z)=(G_1(z),G_2(z))$ that reproduce (to within a certain tolerance) the $x_1(z)$ and $x_2(z)$ evolutions given in \eqref{evolution_1}-\eqref{evolution_2} and thus satisfy the constraint \eqref{ode_constraint} (again to within a tolerance). 

For the numerical experiments we have used $M=10^4$ samples (M/3 used for training, M/3 for validation and M/3 for testing) and mini-batch SGD was used with a mini-batch size of $m=10^3.$ The relative error criterion for the learning rate choosing scheme is set to $TOL = 3/\sqrt{M/3}=3 /\sqrt{10^4/3} \approx 0.052.$

The constraint residual for the GAN generator output is given by 
\begin{equation} \label{constraint_intra_residual_2}
\epsilon_G(z)=(G_1(z)-R)^2+(G_2(z)-R)^2-1. 
\end{equation}
The constraint residual for the true samples was taken to be $\epsilon_D(x) \sim {\cal{N}} (0,0.01^2/M)$ which corresponds to a Gaussian random variable with mean 0 and standard deviation $0.01/\sqrt{M}=0.01/100.$

\begin{figure}[ht]
   \centering
   \subfigure[]{%
   \includegraphics[width = 7cm]{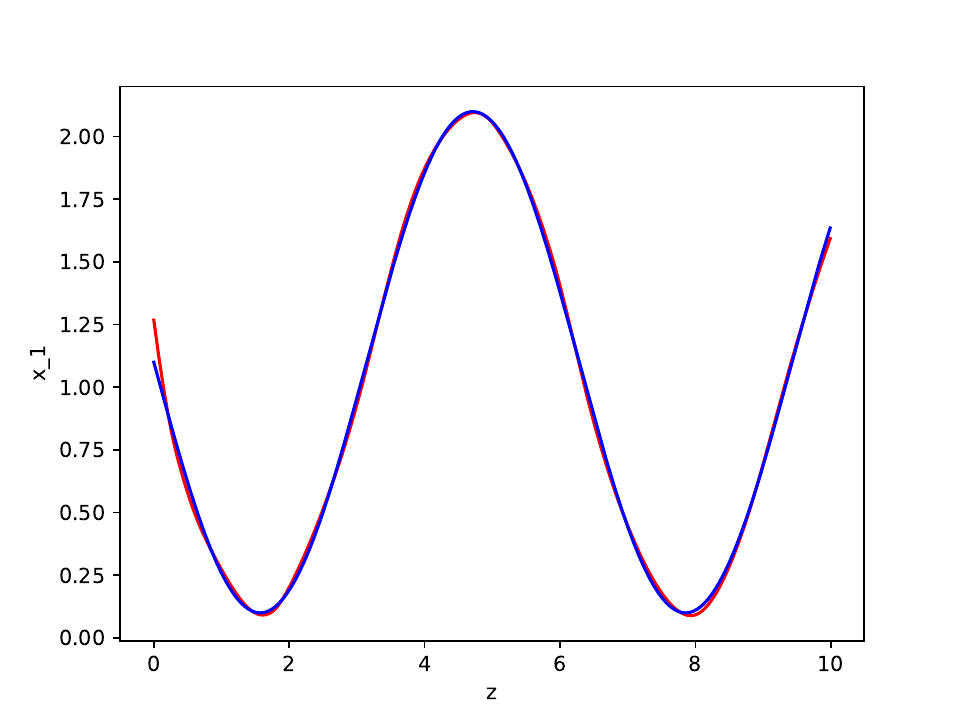}
   \label{plot_harmonic_2a}}
      \quad
   \subfigure[]{%
   \includegraphics[width = 7cm]{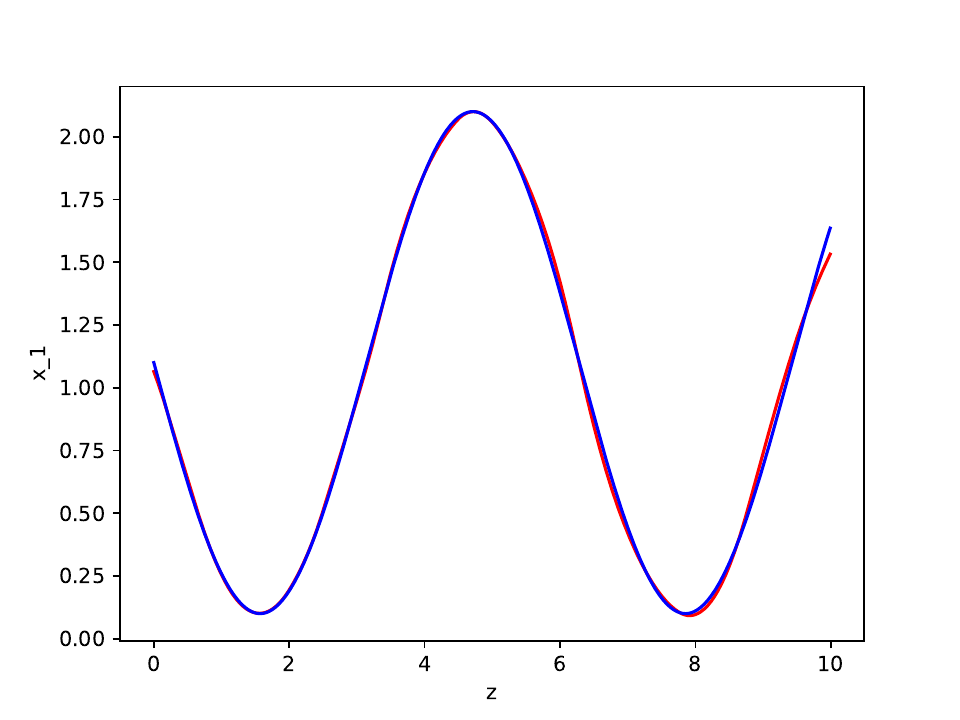}
   \label{plot_harmonic_2b}}
\caption{Comparison of target function $x_1=R -\sin(z) $ (blue line) and actual function learned (red line). (a) $10^4$ samples without enforcing the constraint; (b) $10^4$ samples with the constraint enforced.  }
\label{plot_harmonic_2}
\end{figure} 

\begin{figure}[ht]
   \centering
   \subfigure[]{%
   \includegraphics[width = 7cm]{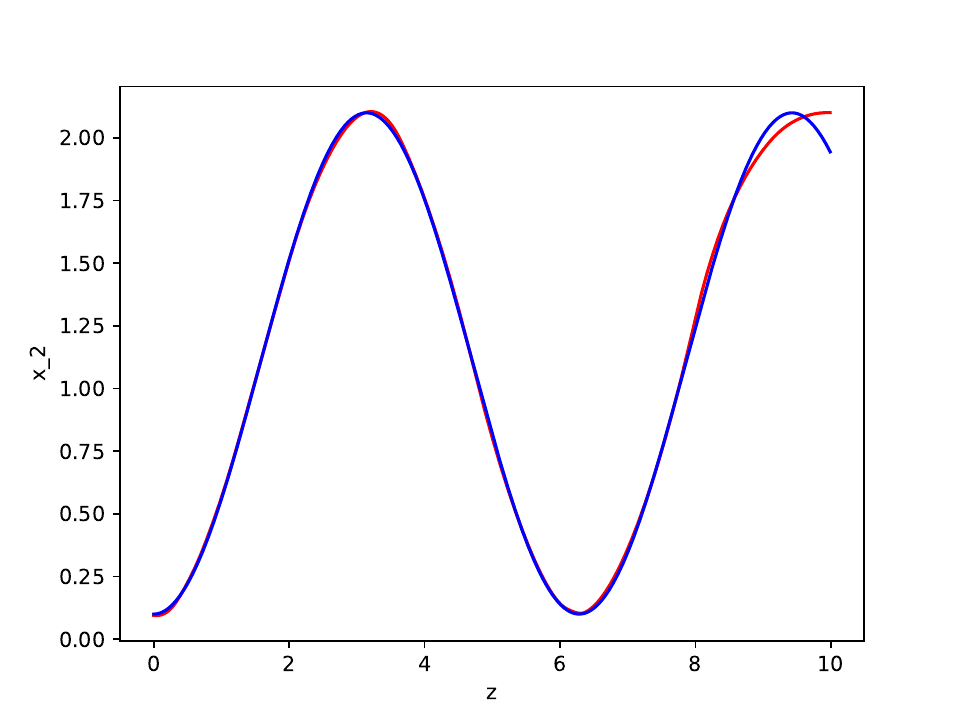}
   \label{plot_harmonic_3a}}
      \quad
   \subfigure[]{%
   \includegraphics[width = 7cm]{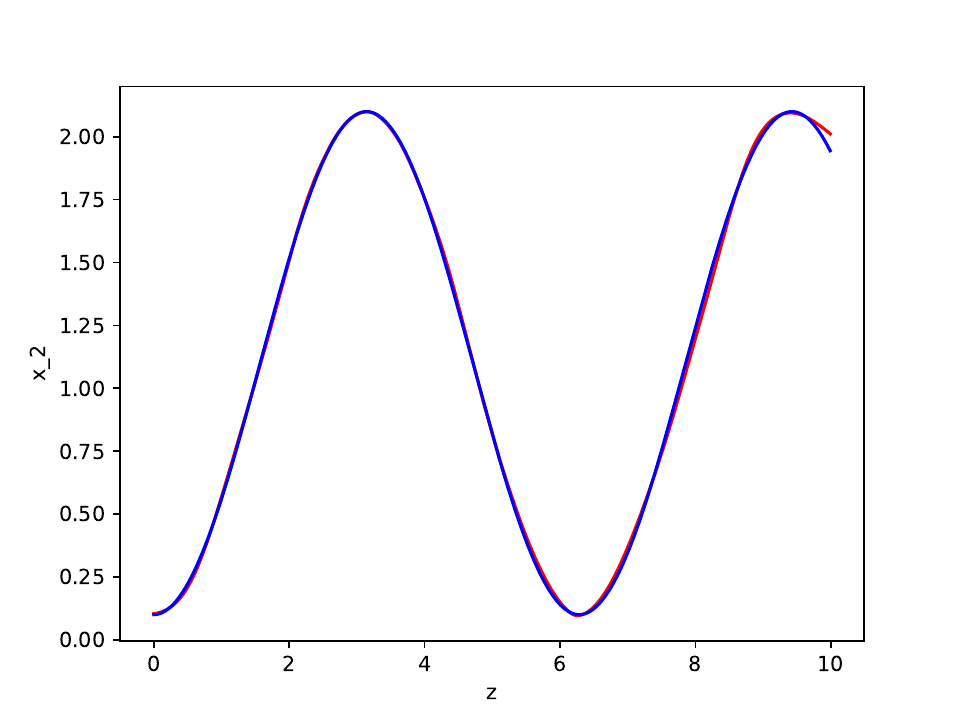}
   \label{plot_harmonic_3b}}
\caption{Comparison of target function $x_2=R -\cos(z) $ (blue line) and actual function learned (red line). (a) $10^4$ samples without enforcing the constraint; (b) $10^4$ samples with the constraint enforced.  }
\label{plot_harmonic_3}
\end{figure}

The relative error used to control the learning rate can be written for this case as
$$ RE_m= \frac{1}{m}\sum_{j=1}^m \biggl[ \frac{|G_1(z_j)-(R-\sin(z_j))|}{|R-\sin(z_j)|} +  \frac{|G_2(z_j)-(R-\cos(z_j))|}{|R-\cos(z_j)|} \biggr] ,$$
where $(G_1(z_j),G_2(z_j))$ is the output of the GAN generator for the input $z_j.$ The parameters for the numerical experiments are $R=1.1$ and $z \in [0,10].$ We note that for the GAN training we allow $z \sim U[0,10].$ Also, we have chosen different deep networks for the generator and discriminator. The generator deep net has 10 hidden layers of width 10 while the discriminator deep net has only 5 hidden layers of width 10. The reason we chose the generator deep net to be larger than that of the discriminator is because this is a more involved case and the discriminator can learn much faster than the generator. As a result, the GAN training often becomes unstable if the discriminator becomes too strict.

\begin{figure}[ht]
   \centering
   \includegraphics[width = 7cm]{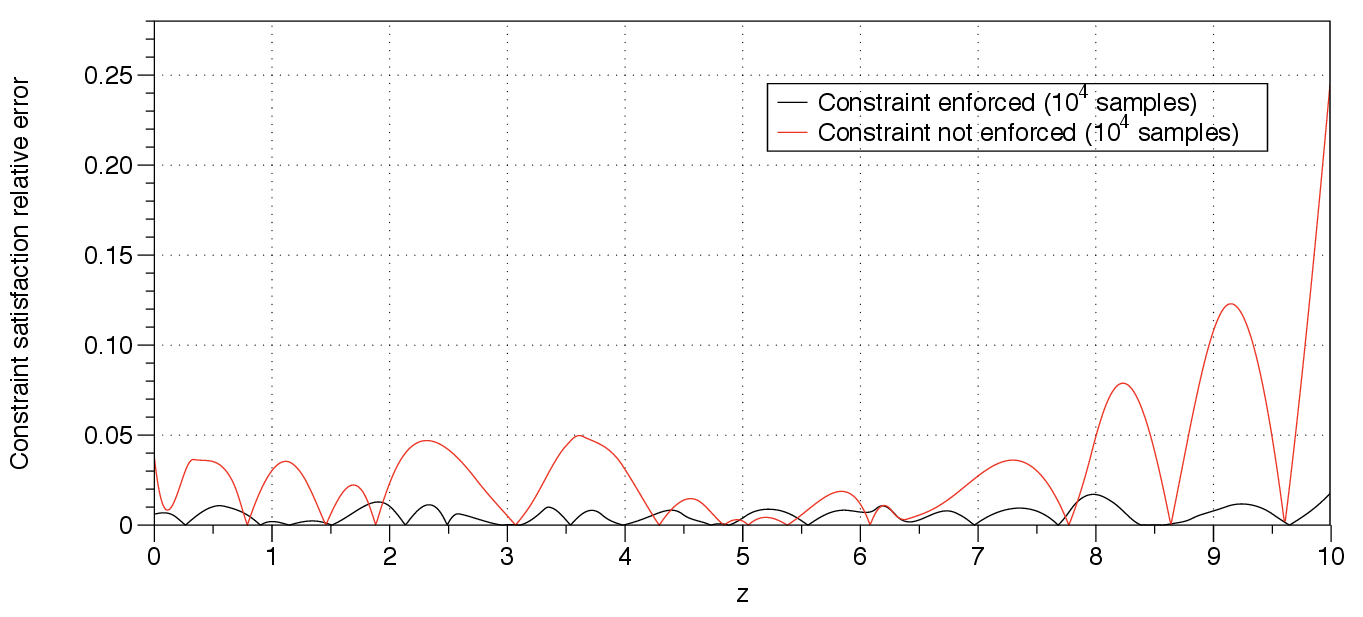}
\caption{Comparison of {\it relative} error in learning the constraint $(x_1(z)-R)^2+(x_2(z)-R)^2=1 .$}
\label{plot_harmonic_3c}
\end{figure}

From Fig. \ref{plot_harmonic_1a} we see that the game-theoretic optimum is reached early on whether we enforce the constraint or not. However, as we can see in Fig. \ref{plot_harmonic_1b} there is a significant difference in the evolution of the relative error. If we enforce the constraint, the GAN's relative error during training converges to within the desired tolerance after 40000 iterations. On the other hand, if we do {\it not} enforce the constraint the relative error does {\it not} converge before the minimum learning rate allowed is reached and the algorithm is terminated. In particular, close to the termination time the relative error hovers around the $6\%$ mark but cannot be reduced further. Recall that according to the termination criterion for convergence discussed at the end of Section \ref{learning_rate}, to establish convergence within $TOL$ we need to have $RE_m \leq TOL$ for at least {\it half} of the $N_{check}=2000$ iterations within an iteration parcel.

Figs. \ref{plot_harmonic_2} and \ref{plot_harmonic_3} contain comparisons of the target function and the actual function learned for $x_1(z)$ and $x_2(z).$ As expected, enforcing the constraint improves the accuracy of the actual function learned. It also results in a dramatic difference in the generator output's relative error of satisfying the constraint given by
\begin{gather*}
RE_m^{constraint}(z) =\frac{|(G_1(z)-R)^2+(G_2(z)-R)^2-\bigl[ (x_1(z)-R)^2+(x_2(z)-R)^2 \bigr]|}{|(x_1(z)-R)^2+(x_2(z)-R)^2|} \\
=|(G_1(z)-R)^2+(G_2(z)-R)^2-1|
\end{gather*}
since $(x_1(z)-R)^2+(x_2(z)-R)^2 =1.$ As we can see in Fig. \ref{plot_harmonic_3c}, if we do {\it not} enforce the constraint, the relative error in the generator output satisfying the constraint is in the range $2\%-25\%.$ On the other hand, if we {\it do} enforce the constraint the relative error $RE_m^{constraint}(z)$ is no more than $2\%$ for {\it all} points. Recall that the relative error $TOL$ that we have used during training is 0.052 i.e. $5.2\%.$ So, enforcing the constraint during training brings, in this example, the relative error in constraint satisfaction to within the same tolerance as that used for training the GAN. This result strengthens our argument that if a constraint is known it should be enforced during training.

\subsection{Enforcing constraints for extrapolation}\label{example_extrapolation}
We turn to the problem of enforcing constraints for extrapolation. In Section \ref{example_extrapolation_linear} we present results for the linear system given by \eqref{ode_1}-\eqref{ode_2}. In Section \ref{example_extrapolation_nonlinear} we present results for the Lorenz system which is nonlinear. For the linear system we examine two cases: i) enforcing constraints only during training and ii) enforcing constraints during training {\it and} during prediction. We do that because the solutions of the system \eqref{ode_1}-\eqref{ode_2} are known to evolve on a circle and thus enforcing this (algebraic) constraint also during prediction is straightforward. For the Lorenz system, we only enforce constraints during training.  

\subsubsection{Linear system}\label{example_extrapolation_linear}

For this example, the task is to train a GAN generator to produce, through iterative application, a trajectory of the system \eqref{ode_1}-\eqref{ode_2} for $z \in [0,10]$ starting from the initial condition $x_1(0)=R$ and $x_2(0)=R-1.$ Because we want to apply the GAN generator iteratively, we need to learn the map of the system i.e., the transformation that takes the state of the system from one instant to the next. For this example, we train the GAN generator to take as input {\it both} the state of the system and the rate of change of the state at one instant (the velocity) and output the state of the system and rate of change of the state at the next instant.

We want to train the GAN generator to produce (through iterative application) the time series
\begin{gather}
x_{pos}^{extra}(z)=x_1(z)=R-\sin(z) \label{extra_evolution_1}\\
x_{rate}^{extra}(z)=\frac{d}{dz} x_1(z)=-\cos(z) \label{extra_evolution_2}\\
y_{pos}^{extra}(z)=x_2(z)=R-\cos(z) \label{extra_evolution_3} \\
y_{rate}^{extra}(z)=\frac{d}{dz} x_2(z)=\sin(z) \label{extra_evolution_4}
\end{gather} 
We have used the notation $x_{pos}^{extra}(z), x_{rate}^{extra}(z), y_{pos}^{extra}(z), y_{rate}^{extra}(z)$ to distinguish the results in this section which involve extrapolation from the results in the previous section where we only performed interpolation. 

\begin{figure}[ht]
   \centering
   \subfigure[]{%
   \includegraphics[width = 7cm]{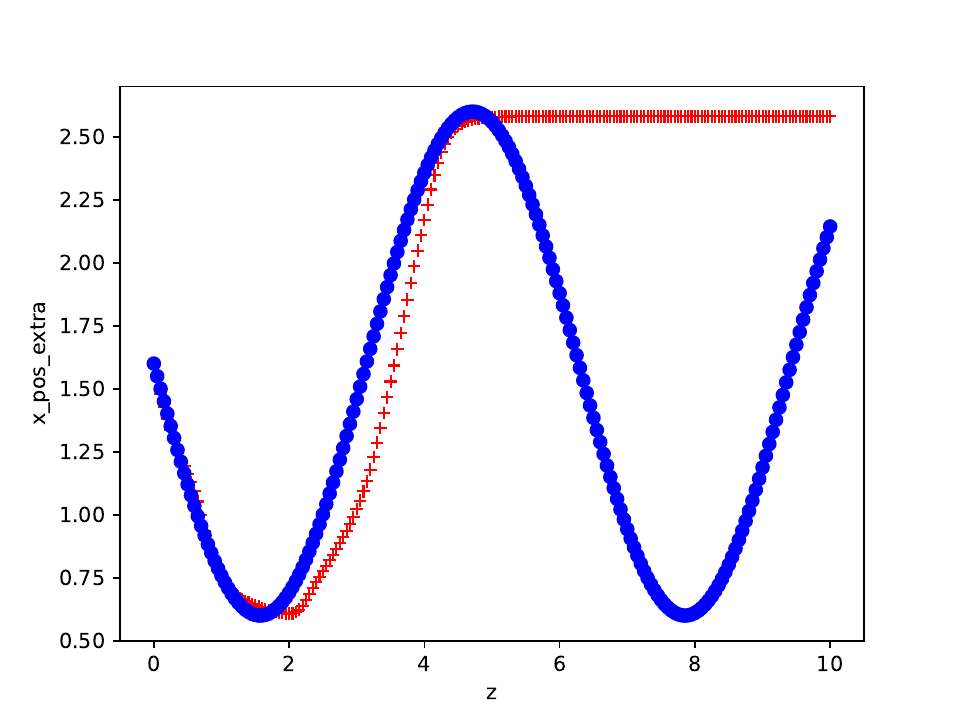}
   \label{plot_harmonic_4a}}
      \quad
   \subfigure[]{%
   \includegraphics[width = 7cm]{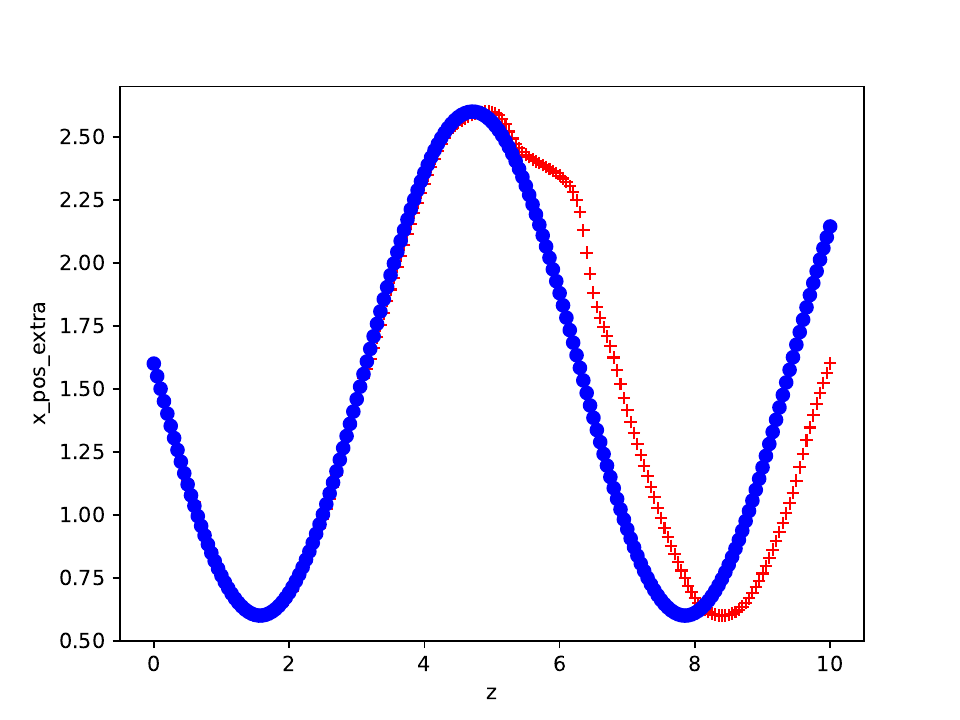}
   \label{plot_harmonic_4b}}
\caption{Comparison of target function $x_1(z)=R -\sin(z) $ (blue dots) and GAN prediction $x_{pos}^{extra}(z)$ (red crosses). (a) $10^4$ samples of {\it noiseless} data {\it with} the constraints enforced during training and {\it no} projection step; (b) $10^4$ samples of {\it noiseless} data {\it with} the constraints enforced during training and a projection step (see text for details).  }
\label{plot_harmonic_4}
\end{figure}

\begin{figure}[ht]
   \centering
   \subfigure[]{%
   \includegraphics[width = 7cm]{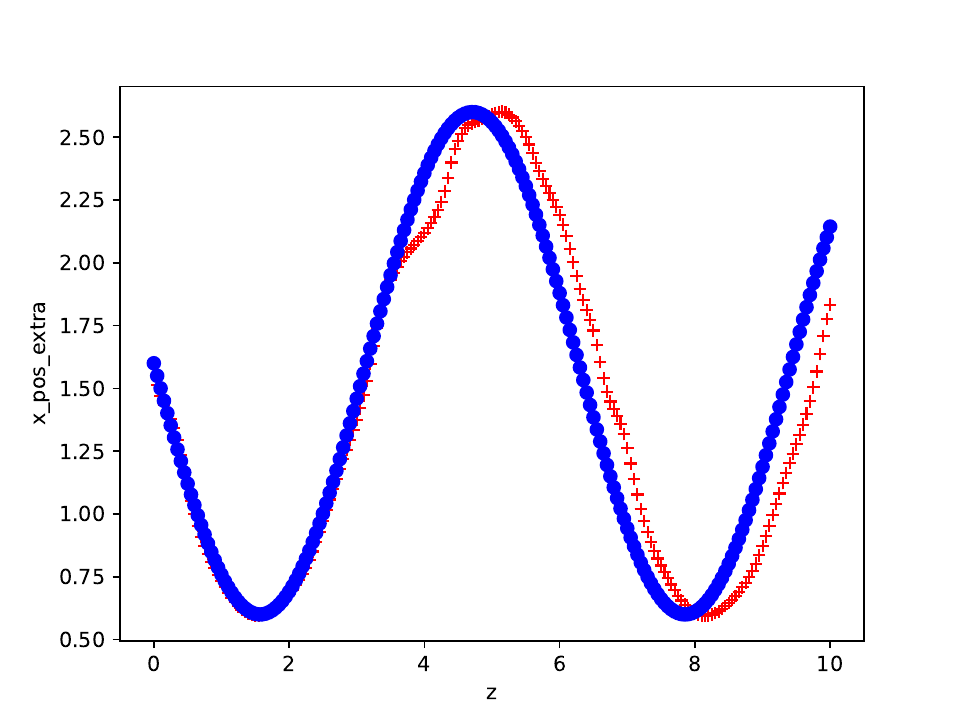}
   \label{plot_harmonic_8a}}
      \quad
   \subfigure[]{%
   \includegraphics[width = 7cm]{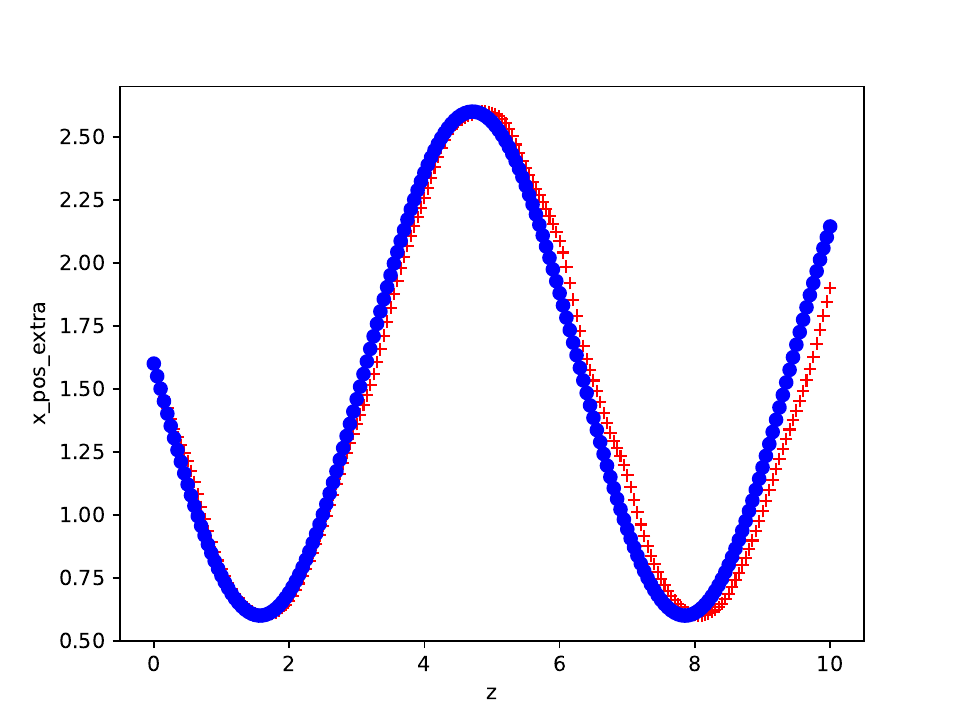}
   \label{plot_harmonic_8b}}
\caption{Comparison of target function $x_1(z)=R -\sin(z) $ (blue dots) and GAN prediction $x_{pos}^{extra}(z)$ (red crosses). (a) $10^4$ samples of {\it noisy} data {\it with} enforced constraints during training and with {\it no} projection step; (b) $10^4$ samples of {\it noisy} data {\it with} enforced constraints during training and with a projection step (see text for details).  }
\label{plot_harmonic_8}
\end{figure}

Figures \ref{plot_harmonic_4} and \ref{plot_harmonic_8} contain the extrapolation result of the GAN generator for the quantity $x_{pos}^{extra}(z)=x_1(z)$ under different scenarios. All the cases used $M=10^4$ samples (with $M/3$ for training, $M/3$ for validation and $M/3$ for testing). Also, the constant $R=1.6.$ Recall that in the previous section, the input of the generator was the random variable $z \sim U[0,10].$

In the previous section, the input was the instant $z.$ In this section, the input of the generator is either the vector $$(x_1(z),\frac{d}{dz} x_1(z),x_2(z),\frac{d}{dz} x_2(z))=(R-\sin(z),-\cos(z), R - \cos(z), \sin(z))$$ for the {\it noiseless} case or $$(x_1(z),\frac{d}{dz} x_1(z),x_2(z),\frac{d}{dz} x_2(z))=(R'-V'\sin(z),-V'\cos(z), R'-V'\cos(z),V'\sin(z))$$ for the {\it noisy} case (see next paragraph for explanation of $R'$ and $V'$). For all scenarios, the output vector is 
\begin{gather*}
(x_1(z + \Delta z),\frac{d}{dz} x_1(z + \Delta z),x_2(z + \Delta z),\frac{d}{dz} x_2(z + \Delta z)) \\
=(R-\sin(z+ \Delta z),-\cos(z + \Delta z),R-\cos(z+ \Delta z),\sin(z + \Delta z)).
\end{gather*}
To further avoid confusion with the notation for the interpolation problem of the previous section, we define the vector
$$x(z)=(x_1(z),\frac{d}{dz} x_1(z),x_2(z),\frac{d}{dz} x_2(z))$$ and also the vector $$G(x(z))=(G_1(x(z)),G_2(x(z)),G_3(x(z)),G_4(x(z)))$$ which is the output of the generator $G$ for input vector $x(z).$

Notice that while we allow in some cases the input to be noisy, the output is {\it always} the point on the exact trajectory. As we have explained in Section \ref{enforcing_constraints_extrapolation}, the use as input of a cloud of points distributed around the exact trajectory and the use of the exact point on the trajectory as the output is a way to incorporate a restoring force in the dynamics learned by the GAN generator.

The random variables $R' \sim U[R-R_{range},R+R_{range}]$ and $V' \sim U[1-V_{range},1+V_{range}]$ determine the cloud of noisy data around each input point for the {\it noisy} case. For the numerical experiments presented in Fig. \ref{plot_harmonic_8} we have used $R_{range}/R= 3 \times 10^{-2}$  and $V_{range}=3 \times 10^{-2}.$ The choice for $R_{range}$ and $V_{range}$ was approximately $1/\sqrt{m}=1/\sqrt{10^3}$  where $m$ is the mini-batch size used in the GAN training. 

We need to comment on the values of the range of the cloud of points as well as the number of points used in the cloud. Ideally, we would like to extract some (scaling) laws which connect the range of the cloud points, say $R_{range}$ or $V_{range},$ with the stepsize $\Delta z.$ This can allow the calibration of the generator GAN in a certain regime and the use of it in other regimes where calibration could be expensive e.g. if we want to use many more samples. The determination of the extent of $R_{range}$ or $V_{range}$ is related to the amplitude of the restoring force that the cloud of points is introducing in the dynamics described by the generator. As mentioned already in Section \ref{enforcing_constraints_extrapolation}, the concept of a restoring force has connections with model reduction \cite{chorinstinis2007}. This subject is currently under investigation and results will be reported in a future publication.

A related question is how many points $N_{cloud}$ one should use to obtain an adequate resolution of $R_{range}$ and $V_{range}.$ We have chosen for the experiments $N_{cloud}=50.$ This choice means that there are $10^4/50=200$ different values of $z \sim U[0,10]$ and the stepsize $\Delta z = 10/200 =5 \times 10^{-2}.$ For the adaptive learning rate schedule we used $TOL = 1/\sqrt{10^4/3} \approx 0.0173.$ We have chosen different deep networks for the generator and discriminator. The generator deep net has 15 hidden layers of width 20 while the discriminator deep net has only 5 hidden layers of width 20.

The relative error used to control the learning rate can be written for this case as
\begin{gather*}
RE_m= \frac{1}{m}\sum_{j=1}^m \frac{1}{4} \biggl[ \frac{|G_1(x(z_j))-(R-\sin(z_j+ \Delta z))|}{|R-\sin(z_j + \Delta z)|} +  \frac{|G_2(x(z_j))-\cos(z_j+ \Delta z)|}{|\cos(z_j+ \Delta z)|} \\
+\frac{|G_3(x(z_j))-(R-\cos(z_j+ \Delta z))|}{|R-\cos(z_j+ \Delta z)|} +  \frac{|G_4(x(z_j))-\sin(z_j+ \Delta z)|}{|\sin(z_j+ \Delta z)|} \biggr] ,
\end{gather*}
where $G(x(z_j))=(G_1(x(z_j)),G_2(x(z_j)),G_3(x(z_j)),G_4(x(z_j)))$ is the output of the GAN generator at $z_j + \Delta z$ for the input vector $x(z_j)$ at time $z_j.$ We have chosen this notation for the generator output to cover both cases of noiseless and noisy data.

The variables satisfy the constraints 
\begin{equation}
(x_1(z)-R)^2+(x_2(z)-R)^2 =1 \label{constraint_extra_1}
\end{equation} and 
\begin{equation}
(\frac{d}{dz} x_1(z))^2+(\frac{d}{dz} x_2(z))^2 =1 \label{constraint_extra_2} 
\end{equation}
for all $z.$ 

This means that during training, for each time $z,$ the constraint residuals for the GAN generator output are
\begin{equation}
\epsilon_{G1}(z)=(G_1(x(z))-R)^2+(G_3(x(z))-R)^2 -1 \label{constraint_extra_1_residual}
\end{equation} and 
\begin{equation}
\epsilon_{G2}(z)=(G_2(x(z)))^2+(G_4(x(z)))^2 -1 \label{constraint_extra_2_residual}. 
\end{equation}
The constraint residual for the true samples  $\epsilon_D(x) \sim {\cal{N}} (0,0.01^2/M)$ which corresponds to a Gaussian random variable with mean 0 and standard deviation $0.01/\sqrt{M}=0.01/100.$

We need to specify also the projection on the space of constraints during prediction. The form of the constraints \eqref{constraint_extra_1} and \eqref{constraint_extra_2} makes the projection step straightforward. In particular, let $(x_{pos}^{extra}(z), x_{rate}^{extra}(z), y_{pos}^{extra}(z), y_{rate}^{extra}(z))$ be the GAN generator output during prediction. Then, the constraints mean that the pairs $(x_{pos}^{extra}(z)-R , y_{pos}^{extra}(z)-R)$ and $(x_{rate}^{extra}(z), y_{rate}^{extra}(z))$ lie on the unit radius circle for all $z.$  To satisfy these requirements we just need to divide $(x_{pos}^{extra}(z)-R , y_{pos}^{extra}(z)-R)$ by $((x_{pos}^{extra}(z)-R)^2 + (y_{pos}^{extra}(z)-R)^2)^{1/2}$ and divide $(x_{rate}^{extra}(z), y_{rate}^{extra}(z))$ by $(x_{rate}^{extra}(z)^2 + y_{rate}^{extra}(z)^2)^{1/2}$ (see also \cite{hairer2001} for more on integrating ordinary differential equations on manifolds). 

Fig. \ref{plot_harmonic_4} shows the extrapolation result when the training of the GAN generator is performed with {\it noiseless} data with the constraints enforced during training and without a projection step (Fig. \ref{plot_harmonic_4a}) or with a projection step (Fig. \ref{plot_harmonic_4b}) added during extrapolation. From these figures we see that if the data used for training are noiseless, even employing a projection step during prediction is not enough to produce accurate extrapolation results.

In Fig. \ref{plot_harmonic_8}, we present extrapolation results when the training of the GAN generator is performed with {\it noisy} data. In Fig. \ref{plot_harmonic_8a}, the constraints are enforced during training but not during extrapolation i.e., no projection step. Compared with Fig. \ref{plot_harmonic_4a} for the noiseless data case, we see that the incorporation of noise in the data improves significantly the extrapolation accuracy of the GAN generator. In Fig. \ref{plot_harmonic_8b}, the constraints are enforced {\it both} during training and during extrapolation. Compared with Fig. \ref{plot_harmonic_4b} for the noiseless data case, we see that the incorporation of noise in the data also improves the extrapolation accuracy of the GAN generator.

As evidenced by Figs. \ref{plot_harmonic_8a} and \ref{plot_harmonic_8b}, for this particular example, the improvement due to the projection step is not very significant. This is because the result without a projection step (Fig. \ref{plot_harmonic_8a}) is already almost in phase with the ground truth. This result is encouraging since the incorporation of a projection step for a general system will in general lead to an optimization problem which can be costly to solve even when initialized from a good starting point.

As we have already commented at the end of Section \ref{enforcing_constraints_extrapolation}, the enforcing of constraints during training is different from the enforcing during extrapolation. During training, the enforcing of constraints involves tuning the parameters of the GAN generator to produce samples that satisfy the constraints. This is achieved through computing constraint residuals for the samples produced from the GAN generator as well as the true samples and augmenting the input vector of the discriminator by these residuals (with some added noise to suppress the GAN instability). During extrapolation, the enforcing of constraints is implemented as a projection step and requires to {\it find a solution that satisfies the constraints}. This will lead in general to an optimization problem which however has a good starting point, the output of the GAN generator before the projection step is applied.

\begin{figure}[ht]
   \centering
   \subfigure[]{%
   \includegraphics[width = 7cm]{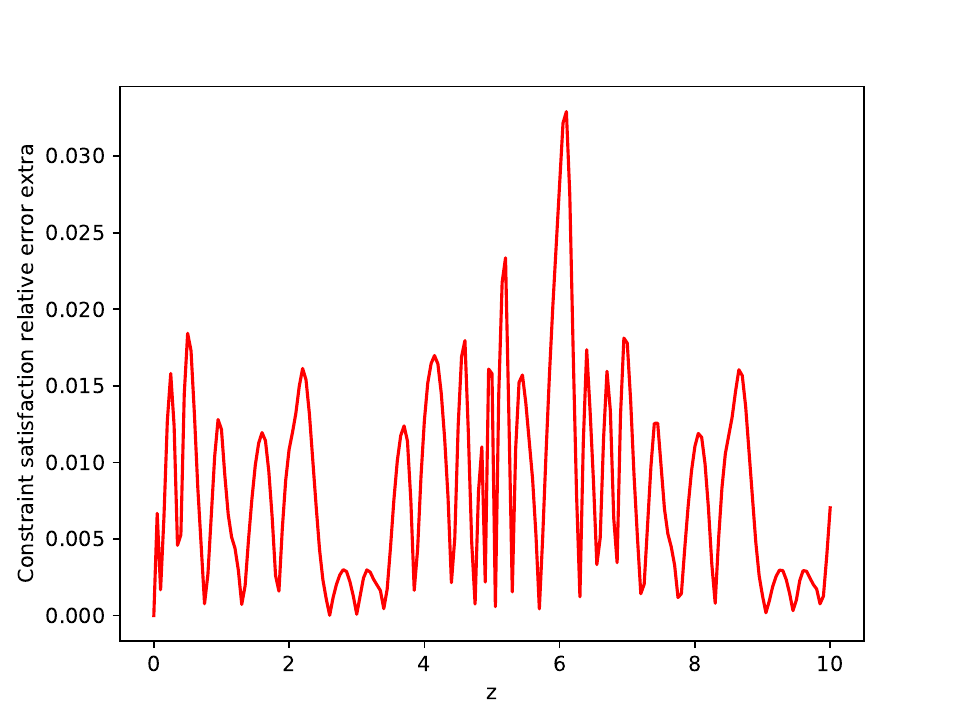}
   \label{plot_harmonic_9a}}
      \quad
   \subfigure[]{%
   \includegraphics[width = 7cm]{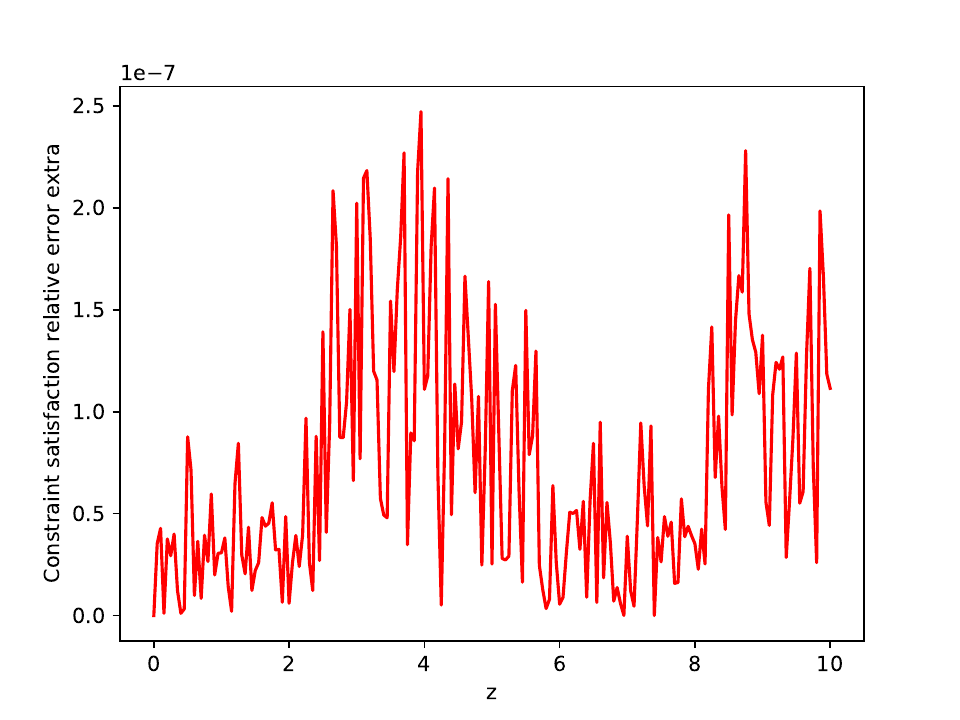}
   \label{plot_harmonic_9b}}
\caption{Relative error in the satisfaction of the constraint $(x_1(z)-R)^2+(x_2(z)-R)^2 =1$ during prediction. (a) $10^4$ samples of {\it noisy} data {\it with} enforced constraints during training and with {\it no} projection step; (b) $10^4$ samples of {\it noisy} data {\it with} enforced constraints during training and with a projection step (see text for details).  }
\label{plot_harmonic_9}
\end{figure}

In Fig. \ref{plot_harmonic_9} we plot the relative error in the satisfaction of the constraint $(x_1(z)-R)^2+(x_2(z)-R)^2 =1$ during prediction. It is given by
\begin{gather*}
RE_m^{constraint}(z) =\frac{|(G_1(x(z))-R)^2+(G_3(x(z))-R)^2-\bigl[ (x_1(z)-R)^2+(x_2(z)-R)^2 \bigr]|}{|(x_1(z)-R)^2+(x_2(z)-R)^2|} \\
=|(G_1(x(z))-R)^2+(G_3(x(z))-R)^2-1|
\end{gather*}
since $(x_1(z)-R)^2+(x_2(z)-R)^2 =1.$ 
As we can see, for the case of prediction {\it with} a projection step, the constraint is satisfied down to numerical accuracy. This is to be expected since the projection step is straightforward to apply exactly on the generator output. However, even for the case of prediction {\it without} the projection step, the relative error in constraint satisfaction never exceeds about $3\%.$

\subsubsection{Nonlinear system}\label{example_extrapolation_nonlinear}

For the last example we have chosen a system for which we do not have a manifold constraint to satisfy at every step. Thus, we will {\it not} use a projection step for the trained GAN generator. Yet, the use of noisy data and the enforcing of constraints during training will allows us to obtain accurate extrapolation results. Unlike the previous examples, here the input and output of the GAN generator is {\it only} the state of the system. In addition, we do not possess an analytical solution but only one produced by a numerical method (the Euler scheme in our case). As will be seen, this permits the {\it explicit} appearance of the restoring force in the constraints to be enforced (see Eqs. \eqref{lorenz_modified1}-\eqref{lorenz_modified2}). The strength of the restoring force is part of the training process and can be adjusted to improve both the GAN generator training and its performance as an extrapolator (predictor).    

We have chosen the Lorenz system
\begin{align}
\frac{d x_1}{dz}&=\sigma (x_2-x_1)  \label{lorenz1} \\
\frac{d x_2}{dz}&= \rho x_1 - x_2 - x_1 x_3  \label{lorenz2}  \\
\frac{d x_3}{dz}&= x_1 x_2 - \beta x_3 \ \label{lorenz3}
\end{align}
where $\sigma, \rho$ and $\beta$ are positive. We have chosen for the numerical experiments the commonly used values $\sigma=10,$ $\rho=28$ and $\beta=8/3.$ For these values of the parameters the Lorenz system is chaotic and possesses an attractor for almost all initial points. We have chosen the initial condition $x_1(0)=0,$ $x_2(0)=1$ and $x_3(0)=0.$ 

Unlike the case of the two coupled linear oscillators in Section \ref{example_extrapolation_linear}, we do not have an analytical expression for the solution of the Lorenz system. We have used as training data the trajectory that starts from the specified initial condition and is computed by the Euler scheme with stepsize $\delta z=10^{-4}.$ In particular, we have used data from a trajectory for $z \in [0,3].$ After we trained the GAN generator we used it to predict the solution for $z \in [0,9].$ This is a severe test of the GAN generator's predictive abilities for three reasons. First, due to the chaotic nature of the Lorenz system there is no guarantee that the GAN generator can correct its errors so that it can follow closely the ground truth trajectory. Second, by extending the interval of prediction beyond the one used for training we want to check whether the GAN generator has actually learned the map of the Lorenz system and not just overfitting the training data. Third, we have chosen an initial condition that is far away from the attractor but our integration interval is long enough so that the system does reach the attractor and then evolves on it. In other words, we want the GAN generator to learn both the evolution of the transient and the evolution on the attractor.  

We performed experiments with different values for the various parameters that enter in our constructions. We present here indicative results for the case of $M=2\times10^4$ samples ($M/3$ for training, $M/3$ for validation and $M/3$ for testing). This gives for the tolerance $TOL = 1/\sqrt{20^4/3} \approx 0.0122.$ The noisy input data for the generator at a point $z$ were given by $x_i'(z)=x_i(z) (1-R_{range}+2R_{range} \times \xi)$ for $i=1,2,3,$ where $\xi \sim U[0,1]$ and $R_{range} =2\times 10^{-2}$ and $x_1(z),x_2(z)$ and $x_3(z)$ is a point on the ground truth trajectory. As in the previous cases, the output of the generator was again {\it noiseless} i.e. $x_i'(z + \Delta z)=x_i(z + \Delta z)$ for $i=1,2,3,$ where $\Delta z$ is the stepsize of the GAN generator. We have chosen $N_{cloud}=100$ for the cloud of points around each input. Thus, the stepsize $\Delta z=1.5 \times 10^{-2}.$ This is because there are $20000/100=200$ instants in the interval $[0,3]$ at a distance $\Delta z = 3/200=1.5 \times 10^{-2}$ apart. Now that we have explained $\Delta z$ we can also explain the value of $R_{range}.$ Recall that the ground truth was computed with the Euler scheme which is a first-order scheme. For the interval $\Delta z=1.5 \times 10^{-2}$ we expect the error committed to be of similar magnitude and thus we should accommodate this error by considering a cloud of points within this range. We found that taking $R_{range}$ slightly larger and equal to $2\times 10^{-2}$ helps the accuracy of the training of the generator.   

Since our solution was produced by the Euler scheme we cannot enforce the exact Lorenz system equations as constraints. We have to devise different constraints that are motivated by the Euler scheme. We say motivated because they cannot be the discretized Euler equations either. The reason is that we have used {\it noisy} input data for the GAN generator. Those input data are mapped to the output which we have taken to be the exact, i.e. {\it noiseless} point on the Euler trajectory. So, the GAN generator learns {\it modified} dynamics. In our case, it is learning a modified Euler scheme where the modification takes the form of a restoring force (from the cloud back to exact trajectory). In the current work, we have opted for the simplest possible modified dynamics. We want the GAN generator to learn to output 

\begin{align}
G_1(x(z)) &= x_1(z) + \Delta z [\sigma (x_2(z)-x_1(z)) -\alpha_1  x_1(z)]  \label{lorenz_modified1} \\
G_2(x(z)) &= x_2(z) + \Delta z [\rho x_1(z) - x_2(z) - x_1(z) x_3(z) -\alpha_2  x_2(z)]  \label{lorenz_modified2}  \\
G_3(x(z)) &= x_3(z) + \Delta z [ x_1(z) x_2(z) - \beta x_3(z) -\alpha_3  x_3(z)] \label{lorenz_modified3}
\end{align} 
where $\alpha_1, \alpha_2$ and $\alpha_3$ are parameters to be optimized during training. For our numerical experiments their magnitudes were $O(10^{-2}).$ 

The constraints \eqref{lorenz_modified1}-\eqref{lorenz_modified3} mean that during training, for each time instant $z,$ the constraint residuals for the GAN generator output are 
\begin{align}
\epsilon_{G1}(z) &= G_1(x(z)) -  \biggr\{ x_1(z) + \Delta z [\sigma (x_2(z)-x_1(z)) -\alpha_1  x_1(z)] \biggl \}\label{lorenz_modified1_residual} \\
\epsilon_{G2}(z) &= G_2(x(z)) - \biggr\{ x_2(z) + \Delta z [\rho x_1(z) - x_2(z) - x_1(z) x_3(z) -\alpha_2  x_2(z)] \biggl\}  \label{lorenz_modified2_residual}  \\
\epsilon_{G3}(z) &= G_3(x(z)) - \biggr\{ x_3(z) + \Delta z [ x_1(z) x_2(z) - \beta x_3(z) -\alpha_3  x_3(z)] \biggl\} \label{lorenz_modified3_residual}
\end{align} 
The constraint residual for the true samples was taken to be $\epsilon_D(x) \sim {\cal{N}} (0,(2\delta z)^2)$ which corresponds to a Gaussian random variable with mean 0 and standard deviation $2\delta z = 2 \times 10^{-4}.$ The reason for this choice of the noise magnitude is that the ground truth is produced through the Euler method with stepsize $\delta z = 10^{-4}.$ Since the Euler method has order 1, we assigned a magnitude of the noise for the constraint residual which is comparable to the error of the numerical method.

Eqs. \eqref{lorenz_modified1}-\eqref{lorenz_modified3} are the constraints we enforce during training and we want to make several observations about them. First, we note that we do not have for the Lorenz system a manifold constraint for the solution that we can enforce also during prediction like we had in the case of the coupled oscillators. So, we enforce constraints only during training. Second, the modifications can be thought of as memory terms that are inserted to account for the unresolved activity in timescales that are smaller than $\Delta z$, the GAN generator's prediction stepsize. In other words, these modification terms signify that the constraints we end up enforcing amount to learning a reduced model where the reduction takes place in the {\it temporal} direction. Third, by allowing the constants to be determined during training, we are in essence training a reduced model of the Lorenz dynamics which is suitable at timescale $\Delta z.$ Four, we have not restricted the sign of the constants to be positive as we would have for true restoring forces. We have found in some numerical experiments that small and negative values can also result in accurate training of the GAN generator. The most probable reason is that the reduced model represented by Eqs. \eqref{lorenz_modified1}-\eqref{lorenz_modified3} is not complete. More elaborate terms are needed to account for all the memory effects. Thus, the assignment of small but negative values signifies that the reduced model is doing its best to accommodate the true reduced dynamics within the bounds of the available functional form. Drawing on our prior experience with model reduction we will investigate the use of more complex reduced models in future work.     

The relative error
\begin{gather*}
RE_m= \frac{1}{m}\sum_{j=1}^m \frac{1}{3} \biggl[ \frac{|G_1(x(z_j))-x_1(z_j+ \Delta z)|}{|x_1(z_j + \Delta z)|} + \frac{|G_2(x(z_j))-x_2(z_j+ \Delta z)|}{|x_2(z_j + \Delta z)|} \\
+\frac{|G_3(x(z_j))-x_3(z_j+ \Delta z)|}{|x_3(z_j + \Delta z)|}  \biggr] ,
\end{gather*}
where $(G_1(x(z_j)),G_2(x(z_j)),G_3(x(z_j)))$ is the output of the GAN generator at $z_j + \Delta z$ for the input vector $x(z_j)$ at time $z_j.$ Also, $(x_1(z_j + \Delta z),x_2(z_j + \Delta z),x_3(z_j + \Delta z))$ is the point on the ground truth trajectory computed by the Euler scheme with $\delta z=10^{-4}.$ For the mini-batch size we have chosen $m=2000.$ We have chosen different deep networks for the generator and the discriminator. The generator deep net has 9 hidden layers of width 20 while the discriminator deep net has only 2 hidden layers of width 20. The numbers of hidden layers both for the generator and the discriminator were chosen as the smallest that allowed the GAN training to reach its game-theoretic optimum without at the same time requiring large scale computations. Also, we note that because the solution of the Lorenz system acquires values outside of the region of the activation function we have removed the activation function from the last layer of the generator.

\begin{figure}[ht]
   \centering
   \subfigure[]{%
   \includegraphics[width = 7cm]{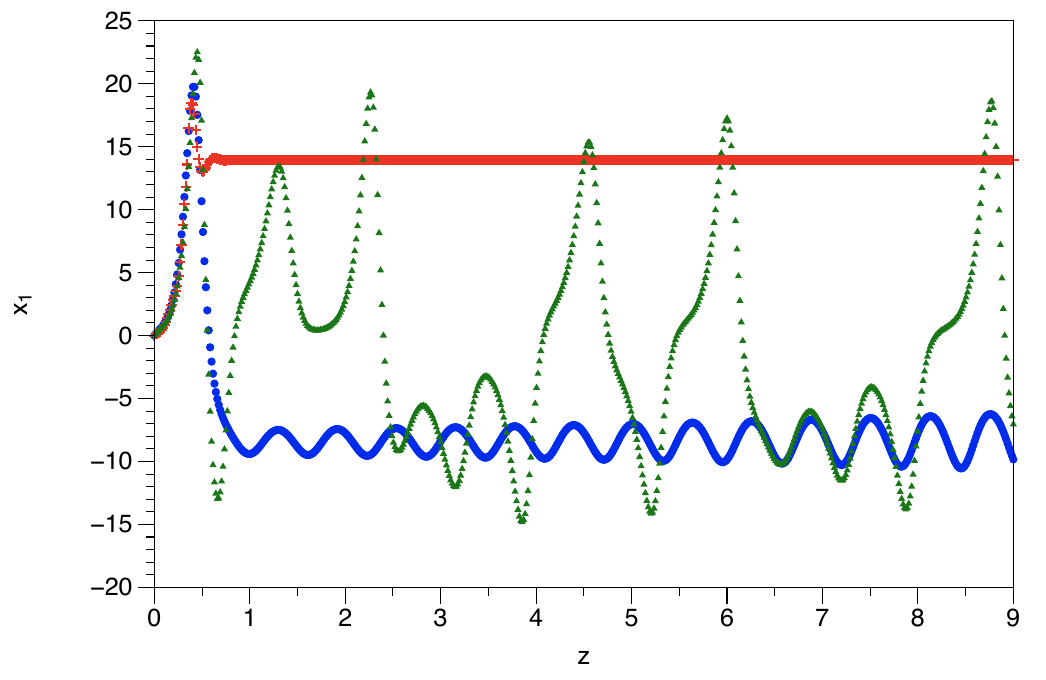}
   \label{plot_lorenz_1a}}
      \quad
   \subfigure[]{%
   \includegraphics[width = 7cm]{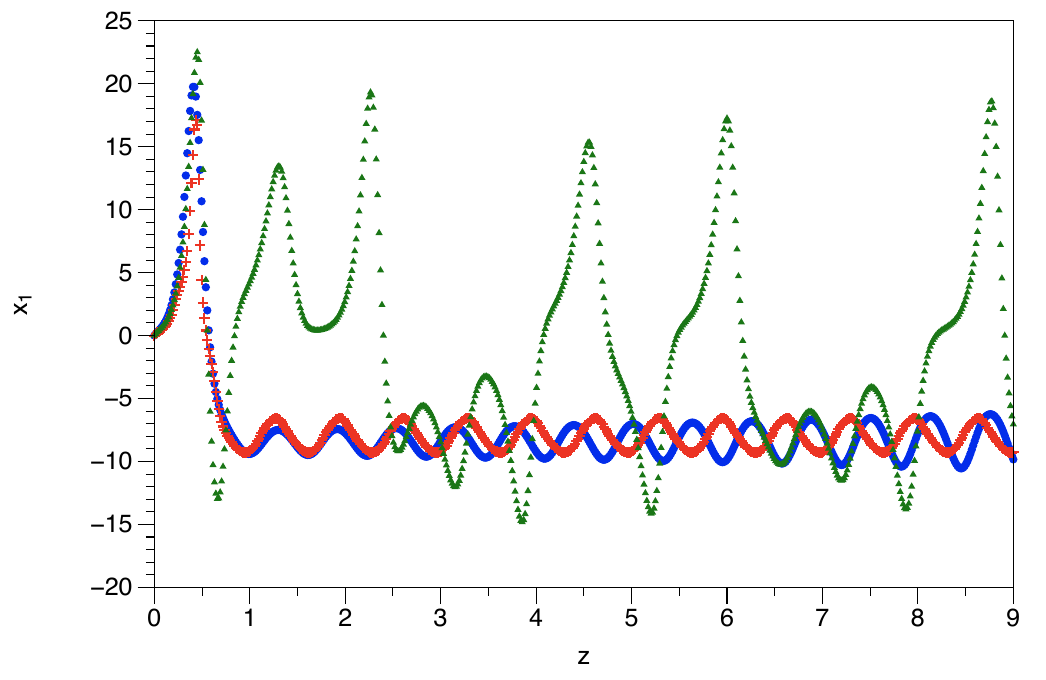}
   \label{plot_lorenz_1b}}
\caption{Comparison of ground truth $x_1(z)$ computed with the Euler scheme with stepsize $\delta z=10^{-4}$(blue dots), the GAN prediction with stepsize $\Delta z=1.5\times10^{-2}$ (red crosses) and the Euler scheme prediction with stepsize $\Delta z=1.5\times10^{-2}$ (green triangles). (a) $2\times 10^4$ samples of {\it noisy} data {\it without enforced constraints} during training; (b) $2\times 10^4$ samples of {\it noisy} data {\it with enforced constraints} during training (see text for details).  }
\label{plot_lorenz_1}
\end{figure}

Fig. \ref{plot_lorenz_1} contains prediction results for the evolution of $x_1(z)$ as predicted by the Euler scheme with $\delta z =10^{-4}$ which is our ground truth, the GAN generator with $\Delta z=1.5\times10^{-2}$ and the Euler scheme with $\Delta z=1.5\times10^{-2}.$ We have included results both for the case of enforcing and not enforcing the constraints \eqref{lorenz_modified1}-\eqref{lorenz_modified3} during training. For the case of {\it not} enforcing constraints depicted in Fig. \ref{plot_lorenz_1a} we see that the GAN generator is unable to follow the ground truth trajectory after about 60 steps (corresponding roughly to a unit of time). On the other hand, enforcing the constraints improves significantly the accuracy of the GAN generator (see Fig. \ref{plot_lorenz_1b}). While not perfect, the prediction of the GAN generator that enforces the constraints \eqref{lorenz_modified1}-\eqref{lorenz_modified3} appears to be able to follow the ground truth trajectory both during its transition to the attractor and during its evolution on the attractor. As expected, the same behavior is observed for the prediction of the other two state variables $x_2(z),x_3(z)$ of the Lorenz system (figures not shown).

As can also be seen in Fig. \ref{plot_lorenz_1b}, the prediction of the GAN generator with enforced constraints improves on the prediction of the Euler scheme with the same stepsize $\Delta z=1.5\times10^{-2}.$ In other words, what the GAN generator with the constraints \eqref{lorenz_modified1}-\eqref{lorenz_modified3} enforced achieves, is a {\it temporal} model reduction of the original Euler scheme. This means that the GAN generator's prediction can follow closely a prediction that was obtained with the Euler scheme but with a smaller stepsize, in our case $\delta z =10^{-4}$ which is 150 times smaller than $\Delta z.$ Of course, the point that training the GAN generator is much more costly than running the Euler scheme for this example is not lost on us. However, what the current simulations show is the affinity between the training of reliable neural network temporal integrators and the concept of model reduction. This is a promising research direction that we will investigate further.   

\begin{figure}[ht]
   \centering
   \subfigure[]{%
   \includegraphics[width = 7cm]{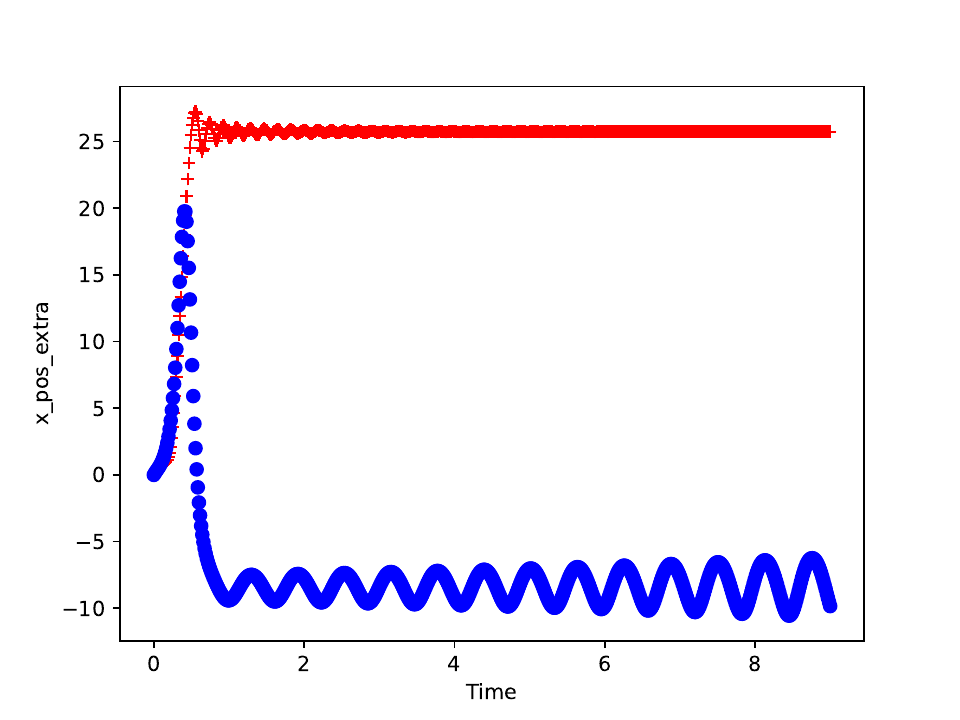}
   \label{plot_lorenz_2a}}
      \quad
   \subfigure[]{%
   \includegraphics[width = 7cm]{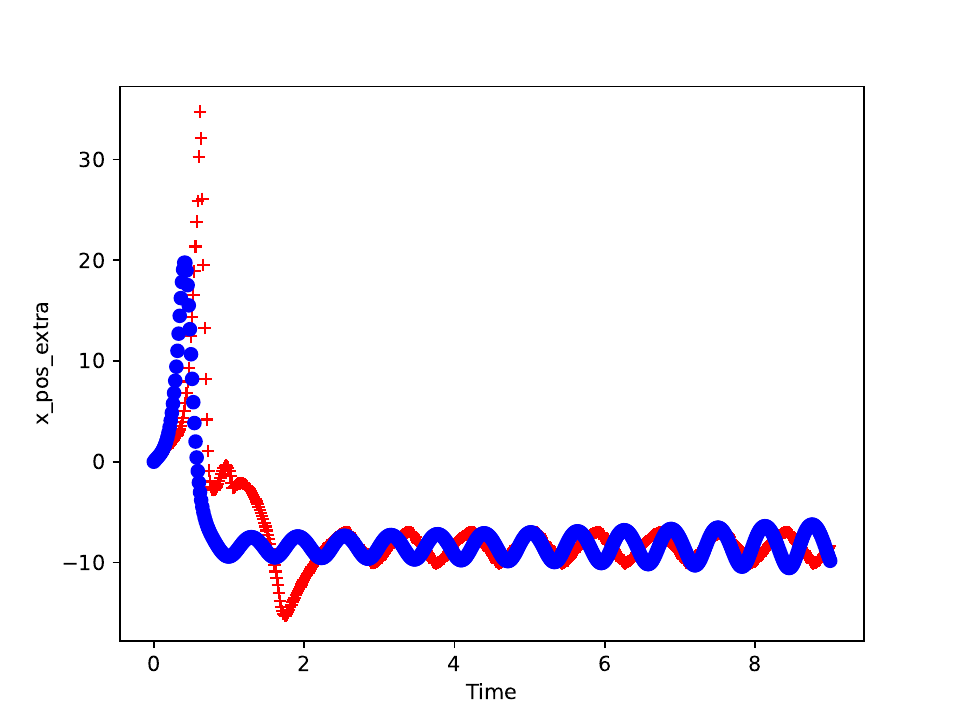}
   \label{plot_lorenz_2b}}
\caption{Comparison of ground truth $x_1(z)$ computed with the Euler scheme with stepsize $\delta z=10^{-4}$(blue dots) and the GAN prediction with stepsize $\Delta z=1.5\times10^{-2}$ (red crosses). (a) $2\times 10^4$ samples of {\it noiseless} data {\it without enforced constraints} during training; (b) $2\times 10^4$ samples of {\it noiseless} data {\it with enforced constraints} during training (see text for details).  }
\label{plot_lorenz_2}
\end{figure}

For comparison purposes to the results of training with noisy data, we show in Fig. \ref{plot_lorenz_2} the prediction results of the GAN generator when it is trained with {\it noiseless} data with and without constraints. For the case of noiseless data, the enforced constraints used did {\it not} include an error-correction term. This is perfectly understandable since the lack of noise in the training data means that we do not have to modify the dynamics. Thus, the constraint residuals were 
\begin{align}
\epsilon_{G1}(z) &= G_1(x(z)) -  \biggr\{ x_1(z) + \Delta z [\sigma (x_2(z)-x_1(z)) ] \biggl \}\label{lorenz_euler1_residual} \\
\epsilon_{G2}(z) &= G_2(x(z)) - \biggr\{ x_2(z) + \Delta z [\rho x_1(z) - x_2(z) - x_1(z) x_3(z)]\biggl\}\label{lorenz_euler2_residual}  \\
\epsilon_{G3}(z) &= G_3(x(z)) - \biggr\{ x_3(z) + \Delta z [ x_1(z) x_2(z) - \beta x_3(z)] \biggl\} \label{lorenz_euler3_residual}
\end{align} 
For the sake of completeness, we did perform a set of experiments with noiseless training data (results not shown) and an error-correcting force and there was no improvement. Also, note that we have used the same number of training samples i.e. we kept $N_{cloud}=100$ but set $R_{range}=0.$ We see that the use of noiseless data for training leads to degradation of the prediction accuracy. This is an encouraging result for our construction given that the production of noisy data is not expensive.

\begin{figure}[ht]
   \centering
   \includegraphics[width = 7cm]{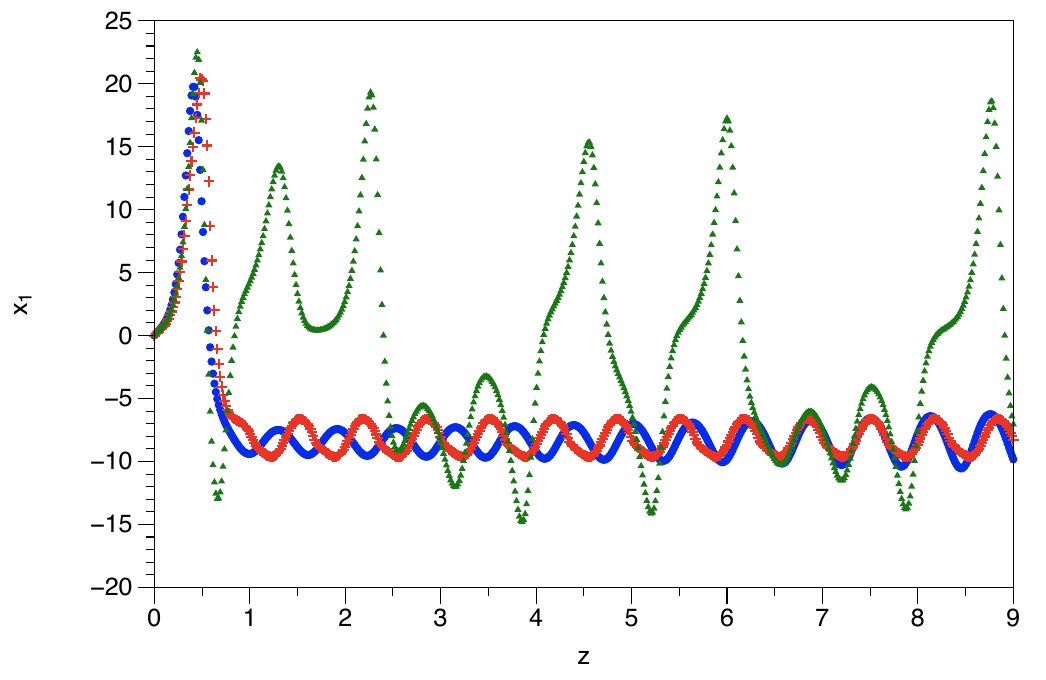}
\caption{Comparison of ground truth $x_1(z)$ computed with the Euler scheme with stepsize $\delta z=10^{-4}$(blue dots), the GAN prediction with stepsize $\Delta z=1.5\times10^{-2}$ (red crosses) and the Euler scheme prediction with stepsize $\Delta z=1.5\times10^{-2}$ (green triangles). During training the discriminator uses as input {\it only} the constraint residuals of a sample (see text for details).  }
\label{plot_lorenz_3}
\end{figure}

Finally, in Fig. \ref{plot_lorenz_3} we present results of a variant of our main construction. Here we choose to use as input for the discriminator {\it only} the constraint residuals for each training sample, omitting any other information about the sample. This means that e.g. for the case of the state variable $x_1(z)$ we feed the discriminator as input only the expression $G_1(z) - x_1(z) - \Delta z [-\sigma (x_2(z)-x_1(z)) -\alpha_1  x_1(z)]$ and not the vector $(x_1(z),G_1(z),G_1(z) - x_1(z) - \Delta z [-\sigma (x_2(z)-x_1(z)) -\alpha_1  x_1(z)]).$ We have kept unchanged all the other parameters of the numerical experiment. We see from Fig. \ref{plot_lorenz_2} that the use of the constraint residual alone for each sample contains enough information for the generator to learn a reasonable approximation of the Lorenz system dynamics. Such a variant of the discriminator input content warrants further investigation since it lowers the dimensionality of the discriminator input and can potentially lead to more efficient training.


\section{Discussion and future work}\label{discussion}
We have presented results of an approach to enforcing constraints for samples generated by a GAN. The purpose of the current work was twofold: i) suggest a way to enforce a constraint that is relatively simple to implement within a given GAN algorithm and ii) exhibit the superior efficiency and/or accuracy afforded by directly enforcing the constraint compared to a GAN algorithm that does not enforce the constraint. We have applied the approach both for the problem of interpolation and extrapolation.

For the problem of interpolation we have shown that the enhancement of the discriminator input vector by the constraint residual (how well the sample created by the generator satisfies the constraint) can improve significantly the efficiency and/or accuracy of a GAN's training. However, in order to reap this benefit one must be careful and respect the fact that {\it the discriminator can learn faster than the generator}. Thus, to avoid instabilities during the GAN training, one has to add a small amount of noise to the constraint residual for the {\it true} data. This allows the generator's training to utilize the extra information about the constraint while keeping the discriminator's training advantage in check.  

For the problem of extrapolation (prediction) of the state of a system, we have shown that a different treatment is needed from the one used for interpolation. In particular, because extrapolation requires the iterative application of the map of the system, using the training data, say from a trajectory, as they are is not enough. We have suggested adding noise to the training data that is used as input to the GAN generator while keeping the output data noiseless. This is akin to adding a restoring force to the system whose evolution we want to compute through extrapolation. The role of the restoring force is to correct the inevitable errors committed by applying iteratively GAN generator. In certain cases, the restoring force can appear explicitly in the enforced constraints. In addition, depending on the problem and the form of the constraints, we can also decide to add a projection step during prediction in order to enforce any given constraints. Such a step is useful for problems where the state of the system is known to be constrained e.g. by a manifold constraint. In this way, our setup resembles a predictor-corrector method, where the GAN generator provides the prediction and the projection step the correction (see \cite{haireretal1987} for more on predictor-corrector methods).

In addition, we have offered a learning rate schedule that bypasses the possible pitfall of the game-theoretic setup where the GAN training has converged but the function learned by the generator is still far away from the target function. To avoid this situation, our learning rate schedule monitors the relative error in learning the target function and/or the constraint. However, our algorithm remains {\it unsupervised} since this information is used to decide only the magnitude of the learning rate and is not back-propagated through the network in order to update the weights. 

We have investigated in the current work how to augment the discriminator input vector in order to enforce constraints in the GAN context. In one of the variants we used {\it only} the constraint residual as an input for the discriminator. We would like to investigate more this and other possible variants that we have not attempted here. In addition, we would like to study the effect of different distributions for the generator inputs on the efficiency/accuracy of the training. Also, all the numerical results presented here involve relatively small networks and were obtained by running the algorithms on a laptop. This is because our aim in the current work was to provide proof of concept about the enforcing of constraints in GANs. We plan to utilize more powerful computational units like GPUs and TPUs when applying the approach to more complex systems.

An interesting research direction to explore is to treat a steady partial differential equation (PDE) with random coefficients (e.g. due to uncertain conductivity or permeability fields) as a constraint. If we have access to solutions from different realizations of the coefficients (for a given set of boundary conditions), we can train a GAN generator to produce sample solutions of the PDE.

Another interesting direction is to study the enforcing of constraints for GANs where the generator is trained to produce the evolution of more complex systems of differential equations. This can be done e.g. by using for the GAN generator a convolutional neural network as was done here or a Long Short Term Memory (LSTM) network  \cite{hochreiterschmidhuber1997}. It can have applications in the construction of integrators for systems that are constrained to evolve on manifolds e.g. the energy surface of a Hamiltonian system or incompressible fluid flows. The idea is that the traditional temporal integrator used in these settings will be replaced by a GAN generator trained with noisy data while the enforcing of the constraints will be incorporated in the training, with a projection step during prediction or both. Such approaches explore {\it how machine learning can aid scientific computing}.   

Another interesting direction of research which shows {\it how scientific computing can aid machine learning}, is to explore the connection between adding noise to the training data and the concept of model reduction. In particular, we want to have a way to decide how large should the added noise be. Since adding noise to the data is analogous to adding a restoring force to the dynamics learned by the generator, we want to have a way to decide how strong should the restoring force be. In the context of model reduction, restoring forces are associated with memory terms \cite{chorinstinis2007}. Thus, we can use this analogy to devise techniques and algorithms through which to estimate the magnitude of the noise that should be added to the data. In the current work, we have explored a simple reduced model for the dynamics of the Lorenz system learned by the GAN generator. We can construct a more elaborate memory term where the memory {\it itself} is represented by a neural network. In this case, there are two neural nets corresponding to the generator that need to be trained simultaneously, one that maps the input to the output and the second that adjusts the memory term during training in order to implement the restoring force. We have been investigating such reduced models and we will report our results in a future publication. We would like to note that noisy data can also be used to train a generator outside of a GAN's {\it unsupervised learning} framework. For example, we can use a traditional {\it supervised learning} setup where one can enforce constraints by adding penalty terms to an objective function. A detailed comparison of the accuracy of the resulting generators for unsupervised and supervised learning will appear elsewhere.  

Such perspectives and constructions can help develop the nascent but fruitful interaction between scientific computing and machine learning.

\section*{Acknowledgements}
The authors would like to thank C. Corley and N. Hodas for very useful discussions and comments. The material presented here is based upon work supported by the Pacific Northwest National Laboratory (PNNL) ``Deep Learning for Scientific Discovery Agile Investment". PNNL is operated by Battelle for the DOE under Contract DE-AC05-76RL01830.

\bibliographystyle{siam}
\bibliography{theory}

\begin{thebibliography}{10}

\bibitem{abadietal2016}
{\sc M.~Abadi, A.~Agarwal, P.~Barham, E.~Brevdo, Z.~Chen, C.~Citro, G.~S.
  Corrado, A.~Davis, J.~Dean, M.~Devin, et~al.}, {\em Tensorflow: Large-scale
  machine learning on heterogeneous distributed systems}, arXiv preprint
  arXiv:1603.04467,  (2016).

\bibitem{arjovskybottou2017}
{\sc M.~Arjovsky and L.~Bottou}, {\em Towards principled methods for training
  {G}enerative {A}dversarial {N}etworks}, arXiv preprint arXiv:1701.04862,
  (2017).

\bibitem{doe_sml_report}
{\sc N.~Baker, F.~Alexander, T.~Bremer, A.~Hagberg, Y.~Kevrekidis, H.~Najm,
  M.~Parashar, A.~Patra, J.~Sethian, S.~Wild, and K.~Willcox}, {\em Workshop
  report on basic research needs for scientific machine learning: Core
  technologies for artificial intelligence},  (2019).

\bibitem{PhysRevE.91.032915}
{\sc T.~Berry, D.~Giannakis, and J.~Harlim}, {\em Nonparametric forecasting of
  low-dimensional dynamical systems}, Phys. Rev. E, 91 (2015), p.~032915.

\bibitem{bu2019}
{\sc Y.~Bu, S.~Zou, and V.~V. Veeravalli}, {\em Tightening mutual information
  based bounds on generalization error}, arXiv preprint arXiv:1901.04609v1,
  (2019).

\bibitem{chen2018}
{\sc R.~T.~Q. Chen, Y.~Rubanova, J.~Bettencourt, and D.~Duvenaud}, {\em Neural
  ordinary differential equations}, arXiv preprint arXiv:1806.07366v3,  (2018).

\bibitem{chorinstinis2007}
{\sc A.~J. Chorin and P.~Stinis}, {\em Problem reduction, renormalization and
  memory}, Communications in Applied Mathematics and Computational Science, 1
  (2007), pp.~1--27.

\bibitem{clevertetal2015}
{\sc D.-A. Clevert, T.~Unterthiner, and S.~Hochreiter}, {\em Fast and accurate
  deep network learning by exponential linear units ({ELU}s)}, arXiv preprint
  arXiv:1511.07289,  (2015).

\bibitem{felsberger2018}
{\sc L.~Felsberger and P.~Koutsourelakis}, {\em Physics-constrained,
  data-driven discovery of coarse-grained dynamics}, arXiv preprint
  arXiv:1802.03824v1,  (2018).

\bibitem{goodfellowetal2014}
{\sc I.~Goodfellow, J.~Pouget-Abadie, M.~Mirza, B.~Xu, D.~Warde-Farley,
  S.~Ozair, A.~Courville, and Y.~Bengio}, {\em Generative adversarial nets},
  Advances in neural information processing systems,  (2014), pp.~2672--2680.

\bibitem{hairer2001}
{\sc E.~Hairer}, {\em Geometric integration of ordinary differential equations
  on manifolds}, BIT, 41 (2001), pp.~996--1007.

\bibitem{haireretal1987}
{\sc E.~Hairer, S.~N\"orsett, and G.~Wanner}, {\em Solving Ordinary
  Differential Equations I}, Springer, NY, 1987.

\bibitem{Han8505}
{\sc J.~Han, A.~Jentzen, and W.~E}, {\em Solving high-dimensional partial
  differential equations using deep learning}, Proceedings of the National
  Academy of Sciences, 115 (2018), pp.~8505--8510.

\bibitem{hochreiterschmidhuber1997}
{\sc S.~Hochreiter and J.~Schmidhuber}, {\em Long short-term memory}, Neural
  Computation, 9 (1997), pp.~1735--1780.

\bibitem{hodasstinis2018}
{\sc N.~Hodas and P.~Stinis}, {\em Doing the impossible: Why neural networks
  can be trained at all}, Frontiers in Psychology,  (2018).

\bibitem{kunze2019}
{\sc J.~Kunze, L.~Kirsch, H.~Ritter, and D.~Barber}, {\em Gaussian mean field
  regularizes by limiting learned information}, arXiv preprint
  arXiv:1902.04340v1,  (2019).

\bibitem{lecunetal2015}
{\sc Y.~LeCun, Y.~Bengio, and G.~Hinton}, {\em Deep {L}earning}, Nature, 521
  (2015), pp.~436--444.

\bibitem{linetal2017}
{\sc H.~W. Lin, M.~Tegmark, and D.~Rolnick}, {\em Why does deep and cheap
  learning work so well?}, Journal of Statistical Physics, 168 (2017),
  pp.~1223--1247.

\bibitem{liu2001}
{\sc J.~S. Liu}, {\em Monte Carlo Strategies in Scientific Computing}, Springer
  NY, 2001.

\bibitem{MaE9994}
{\sc H.~Ma, S.~Leng, K.~Aihara, W.~Lin, and L.~Chen}, {\em Randomly distributed
  embedding making short-term high-dimensional data predictable}, Proceedings
  of the National Academy of Sciences, 115 (2018), pp.~E9994--E10002.

\bibitem{mehtaschwab2014}
{\sc P.~Mehta and D.~Schwab}, {\em An exact mapping between the {V}ariational
  {R}enormalization {G}roup and {D}eep {L}earning}, arXiv: 1410.3831,  (2014).

\bibitem{raissi2018}
{\sc M.~Raissi, P.~Perdikaris, and G.~Karniadakis}, {\em Numerical {G}aussian
  processes for time-dependent and nonlinear partial differential equations},
  SIAM J. Sci. Comput., 40 (2018), pp.~A172--A198.

\bibitem{SIRIGNANO20181339}
{\sc J.~Sirignano and K.~Spiliopoulos}, {\em {DGM}: A deep learning algorithm
  for solving partial differential equations}, Journal of Computational
  Physics, 375 (2018), pp.~1339 -- 1364.

\bibitem{stinis2012}
{\sc P.~Stinis}, {\em Stochastic global optimization as a filtering problem},
  Journal of Computational Physics, 231 (2012), pp.~2002--2014.

\bibitem{wan2018}
{\sc Z.~Wan, P.~Vlachas, P.~Koumoutsakos, and T.~Sapsis}, {\em Data-assisted
  reduced-order modeling of extreme events in complex dynamical systems}, PLoS
  ONE, 13 (2018), p.~e0197704.

\bibitem{xu2017}
{\sc A.~Xu and M.~Raginsky}, {\em Information-theoretic analysis of
  generalization capability of learning algorithms}, in Advances in Neural
  Information Processing Systems 30, I.~Guyon, U.~V. Luxburg, S.~Bengio,
  H.~Wallach, R.~Fergus, S.~Vishwanathan, and R.~Garnett, eds., Curran
  Associates, Inc., 2017, pp.~2524--2533.

\end{thebibliography}

\end{document}